\documentclass[11pt]{article}

\usepackage[preprint]{acl}

\usepackage{times}
\usepackage{latexsym}

\usepackage[T1]{fontenc}

\usepackage[utf8]{inputenc}

\usepackage{hyperref}       
\usepackage{url}            
\usepackage{booktabs}       
\usepackage{amsfonts}       
\usepackage{nicefrac}       
\usepackage{microtype}      
\usepackage{xcolor}         

\usepackage{bm}
\usepackage{hyperref}
\usepackage{url}
\usepackage{booktabs}
\usepackage{multirow}
\usepackage[table]{xcolor} 
\usepackage{siunitx}
\usepackage{graphicx}
\usepackage{tikz}
\usepackage{collcell}
\usepackage{tcolorbox}
\usepackage{float}
\usepackage[dvipsnames]{xcolor}
\usepackage[svgnames]{xcolor}
\usepackage{subcaption}
\usepackage{wrapfig}
\usepackage{xspace}

\usepackage{enumitem}
\usepackage{amsmath, amsthm, amssymb,mathtools} 
\usepackage{cleveref}
\usepackage{array}
\usepackage{arydshln}
\usepackage{tocloft}
\usepackage{titletoc}
\usepackage{natbib}
\usepackage{tabularx}

\newtheorem{definition}{Definition}
\definecolor{lightblue}{RGB}{210,235,255}
\newcommand{\zjc}[1]{\textcolor{blue}{[#1]}}

\newcommand{\direction}{\textsf{TrueThinking}}
\newcommand{\scorename}{\text{TTS}}
\newcommand{\testa}{\text{Engagement Test}}
\newcommand{\testb}{\text{Disengagement Test}}
\newcommand{\faithscore}[1]{$\text{TTS}(#1)$}
\newcommand{\ateNec}{$\mathrm{ATE}_\text{nec}(1)$}
\newcommand{\ateSuf}{$\mathrm{ATE}_\text{suf}(0)$}

\author{
  Jiachen Zhao$^{1}$\thanks{ indicates equal contribution} \thanks{Corresponding author. Email: zhao.jiach@northeastern.edu}\quad
  Yiyou Sun$^{2}$\footnotemark[1] \quad
  Weiyan Shi$^{1}$\thanks{Indicates equal advising role.} \quad
  Dawn Song$^{2}$\footnotemark[3] \\[4pt]
  $^{1}$Northeastern University\\ 
  $^{2}$ University of California, Berkeley\\[3pt]
}

\title{ Can Aha Moments be Fake? Towards Quantifying Decorative and True Thinking in Chain-of-Thought}

\begin{document}
\maketitle

\begin{abstract}
\looseness=-1

Large language models can generate long chain-of-thought (CoT) reasoning, yet prior work suggests that CoT can be post-hoc rationalization rather than a faithful reflection of the computation through explicitly designed settings.  In this work, we go further and propose a \text{True Thinking Score (TTS)} to quantify the causal contribution of each step in CoT to the model's final prediction in realistic reasoning problems. Across eleven models ranging from 1.5B to 1.1T parameters on common reasoning benchmarks, we find that CoTs often interleave \emph{true-thinking} steps, which causally affect the final answer, with \emph{decorative-thinking} steps, which appear useful but have little causal influence; Such decorative steps remain prevalent even for frontier models: Over 30\% of steps in Kimi-K2.6 are decorative on MATH with TTS $\leq$ 0.005. Furthermore, TTS enables effective CoT pruning: removing 50\% of CoT steps with the lowest TTS can largely maintain the performance. Self-training on these pruned CoTs reduces reasoning length by 66\% while preserving performance on Nemotron3-Nano-30B. Finally, we provide a mechanistic analysis showing that LLMs can be steered in the latent space to engage or disengage with reasoning steps. Overall, our results reveal that frontier LLMs often verbalize reasoning steps that are not causally used, challenging both the efficiency and the trustworthiness of CoT\footnote{Our code is released at \href{https://github.com/andotalao24/Identify_true_decorative_thinking}{\url{https://github.com/andotalao24/Identify\_true\_decorative\_thinking}}}. 

\end{abstract}
\vspace{-5mm}

\begin{figure*}[h]
    \centering
    \includegraphics[width=0.88\linewidth]{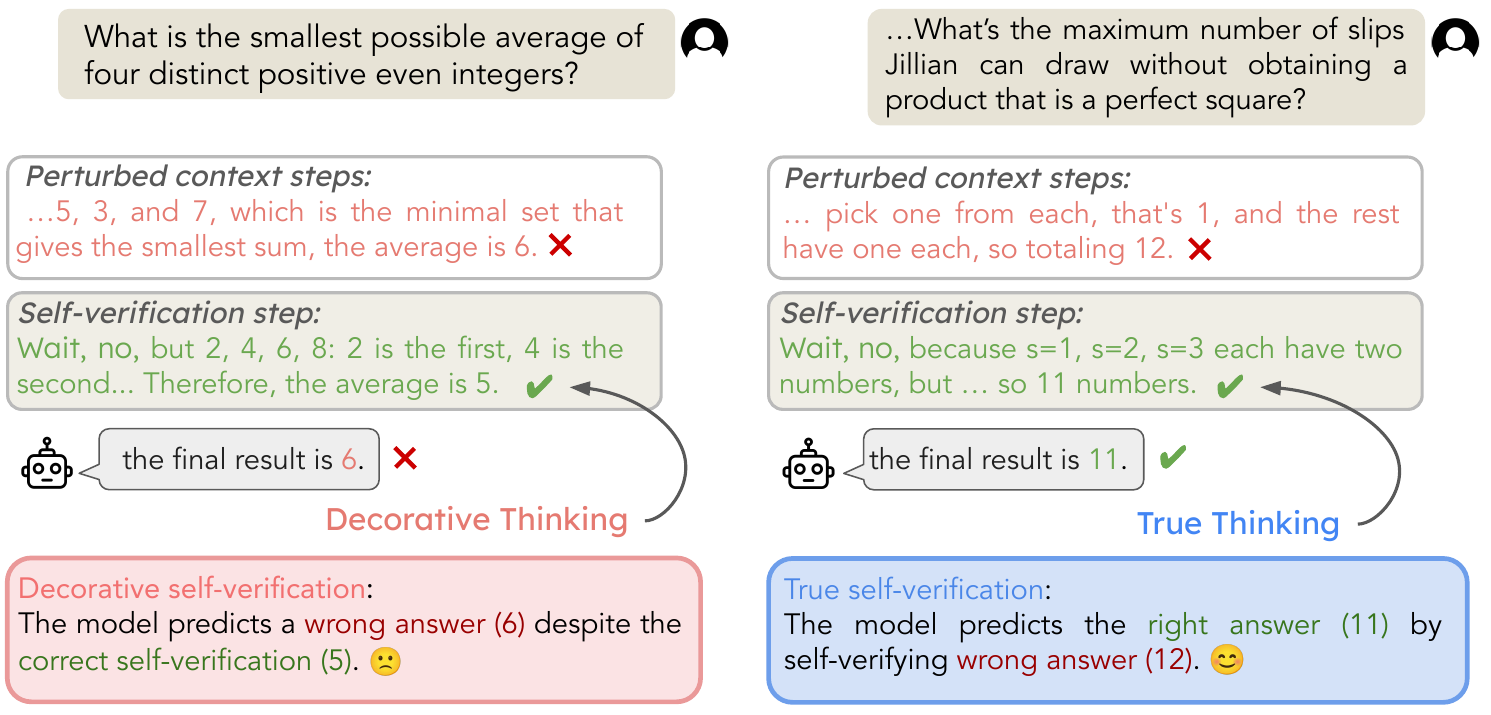}
 
    \caption{\looseness=-1 We find that many CoT steps are not \emph{true thinking}, but act as \emph{decorative thinking}: they contribute little causally to the model's answer. Consider self-verification steps. We randomly perturb numerical values in the preceding context and re-prompt the model to answer using the modified CoT. In the left example, although the self-verification step is correct, the model ignores it and outputs the wrong answer after perturbation. In the right example, the model follows the self-verification step and recovers the correct answer. To quantify this distinction, we propose the True-Thinking Score, which measures the causal contribution of each reasoning step to the model's final prediction. 
    }
    \label{fig:unfaithful-self-check}
\end{figure*}

\section{Introduction}
\looseness=-100
Recent frontier LLMs can increasingly solve complex reasoning problems through test-time scaling, often by generating a very long chain of thought (CoT)~\citep{guo2025deepseek,muennighoff2025s1,snell2024scaling,jaech2024openai}. In their long CoT, these models frequently exhibit an ``aha moment'', where the model begins to \emph{self-verify} its solution~\citep{guo2025deepseek}. LLMs' generated CoT is commonly assumed as a scratch pad where the model thinks out loud~\citep{korbak2025chain}. This has also motivated using CoT as a means to monitor LLMs and detect unsafe behaviors revealed in their CoT~\citep{baker2025monitoring}. 

\looseness=-100
A central assumption behind CoT is {faithfulness}: the verbalized reasoning should reflect the computation that produces the model's final answer. Prior work shows that this assumption can fail. Models may rely on inserted hints~\citep{chen2025reasoning,chua2025deepseek,turpin2023language} or spurious biases~\citep{arcuschin2025chain} without acknowledging them in their CoT, and may commit to an answer before completing the reasoning trace~\citep{ma2025reasoning,wang-etal-2025-chain,yang2025dynamic}. These findings suggest that CoT steps can be post-hoc rationalizations rather than causally used parts of the prediction process.

However, prior work largely relies on controlled settings and remains suggestive, typically showing CoT can be unfaithful at the level of the entire reasoning trace. It still remains unclear how each individual CoT step truly contributes to the model's final prediction in common reasoning tasks. This gap limits the use of CoT as a monitoring signal. If a reasoning step does not causally affect the prediction, inspecting it may give a misleading view of the model's computation, especially in high-stakes settings~\citep{zhong2025impossiblebench,baker2025monitoring}.


\looseness=-1000
To close this gap, we propose the True-Thinking Score (TTS), a step-wise causal analysis of CoT that quantifies how much each verbalized step affects the model’s prediction. We define \textbf{true-thinking} steps as those that causally affect the prediction, and \textbf{decorative-thinking} steps as those that make minimal causal contribution to the prediction, as illustrated in Figure~\ref{fig:unfaithful-self-check}. \scorename{} is adapted from Average Treatment Effect (ATE)~\citep{rubin1974,pearl2009causality} with two complementary interventions by conditioning on its context steps to evaluate its necessity and sufficiency to the model's prediction (Section~\ref{sec:score}).  On common reasoning benchmarks, we evaluate eight frontier models across different scales and find that decorative steps remain widespread (Section~\ref{sec:eval}). On MATH benchmark, nearly 60\% of Qwen3.6-35B's CoT steps have \scorename{} $\leq 0.005$, indicating minimal causal impact on the final answer. Larger models tend to achieve higher \scorename{}, but still contain many decorative steps: for Nemotron3-Super-120B, nearly 30\% of steps are decorative. Moreover, true-thinking and decorative-thinking steps are often interleaved within the same CoT, raising concerns about using CoT as a monitoring signal for model misbehavior~\citep{baker2025monitoring}.

\looseness=-1000
We next show that \scorename{} enables effective CoT pruning (Section~\ref{sec:prune-cot}). Removing the 50\% lowest-\scorename{} steps largely preserves performance, while removing high-\scorename{} steps sharply degrades reasoning accuracy. TTS-based pruning also outperforms an expert prompt-compression LM and direct prompting for concise reasoning. Further self-training on pruned CoTs reduces reasoning length by 66\% while preserving performance, and generalizes to held-out dataset.

\looseness=-1000
Finally, we provide a mechanistic study of step-level causal use (Section~\ref{sec:exp-steer}). We demonstrate that steering the hidden states of a reasoning step can mediate whether the model uses that step to compute the answer. The steering direction is extracted from the contrast between low- and high-\scorename{} steps on one dataset, yet generalizes to other datasets and appears to modulate the model's attention to the targeted step. In particular, steering along the \direction{} direction can induce the model to reason over otherwise decorative self-verification steps and correct previous errors.

Overall, our core contributions are threefold. \begin{itemize}
    \item We propose \scorename{} to quantify the causal contribution of each step in a CoT. 
We reveal that decorative thinking steps---steps with minimal causal contribution---are prevalent in frontier models on common reasoning benchmarks.
\item We show that \scorename{} can be used to prune CoTs without degrading performance, and that further self-training on pruned CoTs improves reasoning efficiency.
\item We provide a mechanistic study, revealing for the first time that whether the model uses a reasoning step to compute the answer can be mediated through steering in the latent space.
\end{itemize}

Together, these findings call for more future work on improving the causality of LLM reasoning, making CoTs more trustworthy and efficient for real-world deployment.

\section{Measuring Step-wise Causality in Reasoning}
\label{sec:framework}
\vspace{-2mm}
\looseness=-1000
In this section, we adapt the Average Treatment Effect (ATE)~\citep{rubin1974,pearl2009causality} to measure the causal impact of individual steps \( s \) on LLMs' prediction result  \( y^* \). We define that a step with genuine causal impact is a \textbf{true-thinking step}, where the model follows \(s\) in order to produce \(y^*\). By contrast, a step with minimal causal impact on \(y^*\) is a \textbf{decorative-thinking step}. We first detail the notations and explain the True-Thinking Score in Section~\ref{sec:score}.

\looseness=-1
\textbf{Notation and Setup.} We adopt notation following \citet{rubin1974,pearl2009causality}.
Let the input question be \( q \), and let the model’s full chain of thought (CoT) for \( q \) be
\(
\mathcal{C}^{\star} = (s_1, s_2, \dots, s_n),
\)
where each \( s_i \) denotes a reasoning step.
At the current step \( s\) under evaluation (we omit the index $i$ and directly use $s$ for simplicity), we define the \textbf{context} as its preceding steps, i.e., 
\(
\mathcal{C} = (s_1, s_2, \dots, s_{i-1})
\). 
To probe the model’s current prediction after any partial reasoning trace, we use early-exit answering by appending a standardized cue: {The final result is}. This approach, following \citet{lanham2023measuring,fu2025deep,yang2025dynamic,tanneru2024hardness,bogdan2025thought}, reliably elicits the model’s intermediate answer given the question \( q \) and reasoning prefix \((\mathcal{C}, s)\).
Let \( f(q, \mathcal{C}, s) \) denote the model’s early-exit prediction after processing \( q \) with context \(\mathcal{C}\) and step \( s \).
The \textit{reference prediction} under the full reasoning trace is then defined as
\(
y^{*} := f(q, \mathcal{C}^{\star}),
\)
representing the model’s final answer when all steps in the full CoT are intact.

\textbf{Indicator Variables.} We introduce the following binary random variables to formalize interventions on reasoning steps:  (1) \textbf{Context indicator} \(\mathbf{C} \in \{0,1\}\):
  \(\mathbf{C}{=}1\) indicates an intact context (the original prefix \(\mathcal{C})\);
  \(\mathbf{C}{=}0\) indicates a perturbed context in which all preceding steps are replaced by perturbed versions.
  We write \( c \in \{0,1\} \) for a specific realization of \(\mathbf{C}\).  (2) \textbf{Step toggle} \(\mathbf{X} \in \{0,1\}\):
  \(\mathbf{X}{=}1\) is the original step \( s \);
  \(\mathbf{X}{=}0\) replaces it with a perturbed version \( s' \). (3) \textbf{Outcome indicator} \(\mathbf{Y} \in \{0,1\}\):
  Given \( f(q, \mathcal{C}, s) \), we define
  \(
  \mathbf{Y} := \mathbf{1} \{ f(q, \mathcal{C}, s) = y^{*} \},
  \)
  which measures whether the model’s early-exit prediction under the given intervention matches the full-CoT reference outcome.

\textbf{Perturbation Procedure.}
To isolate the causal effect of each reasoning step, we create perturbed versions of steps and contexts by introducing \textit{small random numerical offsets} to quantities appearing in the reasoning text \citep{lanham2023measuring,gaoshapley}.
These perturbations are minimal and preserve grammatical and semantic structure, ensuring that the modified step remains coherent but subtly altered.
Additional implementation details are provided in Appendix~\ref{apdx:imp}.

\begin{figure*}[t]
    \centering
    \begin{subfigure}[b]{0.62\linewidth}
        \centering
        \includegraphics[width=\linewidth]{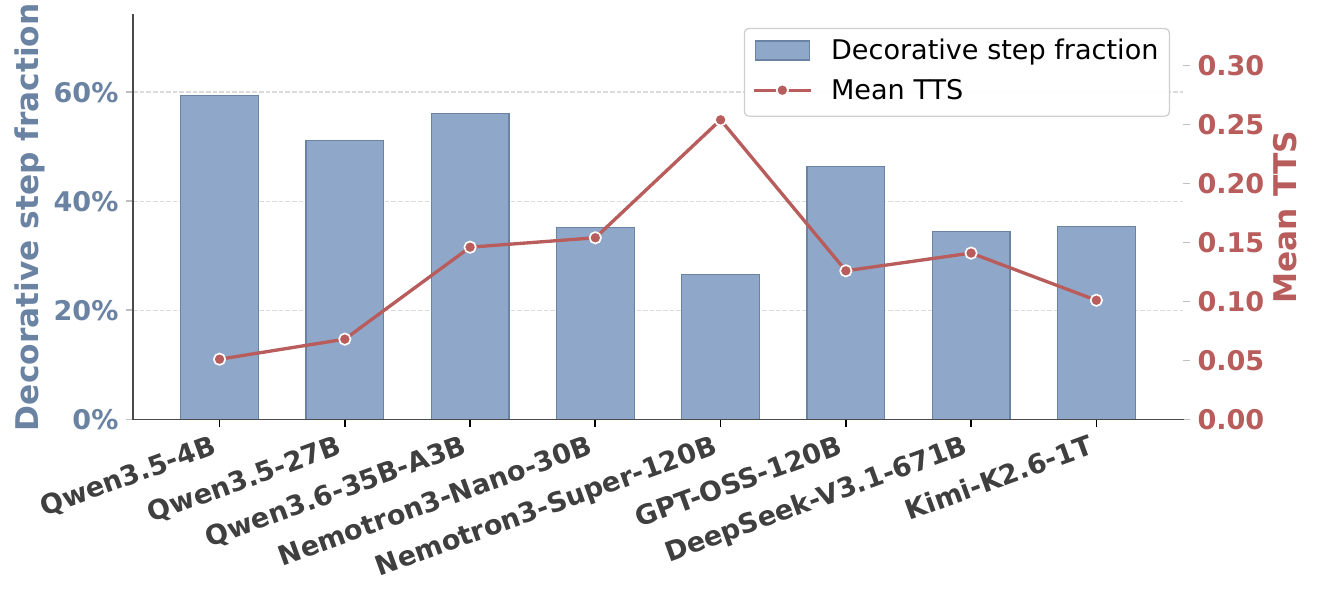}
        \caption{}
        \label{fig:decorative-rate}
    \end{subfigure}
 ~
    \begin{subfigure}[b]{0.36\linewidth}
        \centering
         \includegraphics[width=\linewidth]{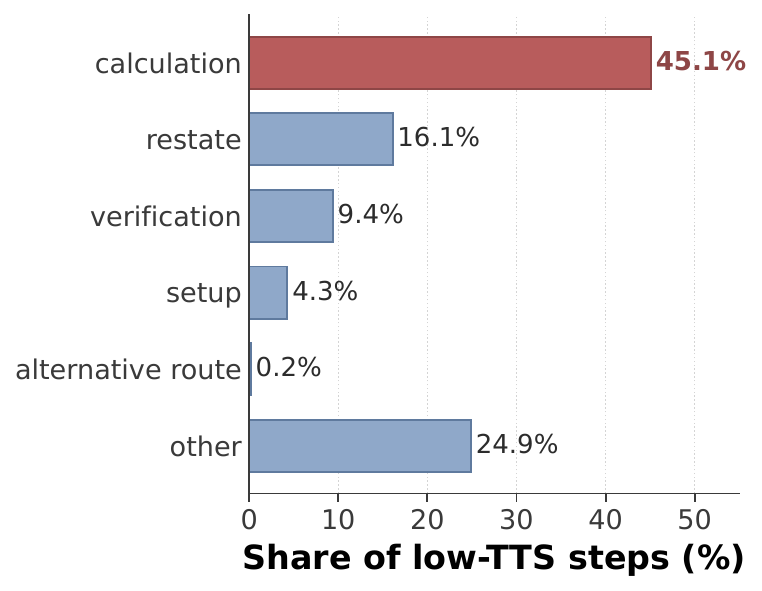}
        \caption{}
        \label{fig:low-tts-types}
    \end{subfigure}
    \caption{(a) Average fraction of steps with \scorename{} $\leq 0.005$ in CoT and average \scorename{} across eight frontier models on MATH. Larger models contain fewer decorative reasoning steps, yet many steps still contribute minimally to the prediction. Our results suggest that the latest models can produce substantially performative and redundant CoT on common reasoning benchmarks, which thus cannot be fully trusted as their true computation process. (b) Corresponding type breakdown of those low-TTS steps. 
}
    \label{fig:decorative-overview}
\end{figure*}

\begin{figure*}[t]
  \centering
  \begin{subfigure}[t]{0.32\textwidth}
    \centering
    \includegraphics[width=\linewidth]{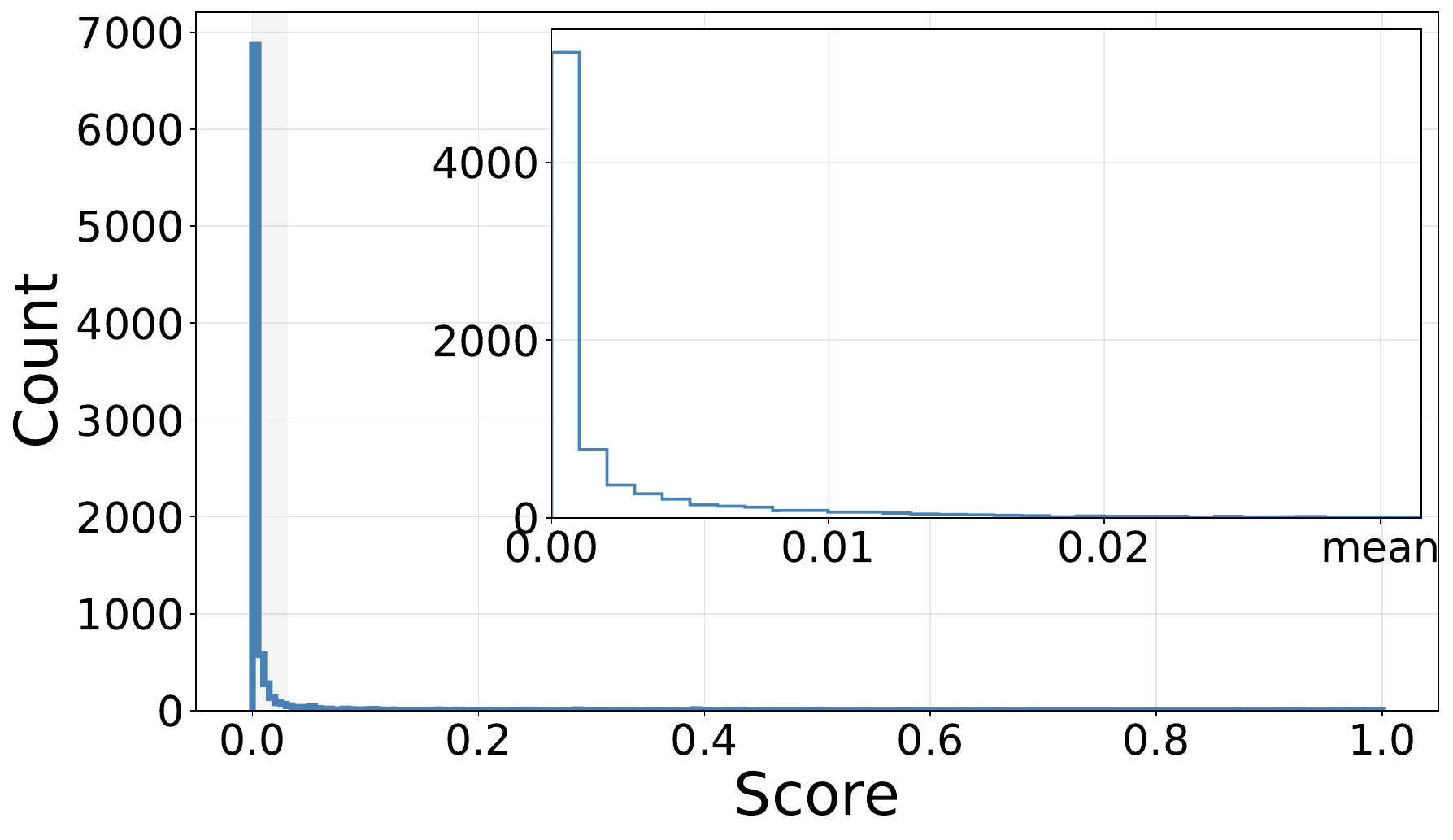}
    \caption{}
    \label{subfig:eval-ret-faith-a}
  \end{subfigure}
  \begin{subfigure}[t]{0.32\textwidth}
    \centering
    \includegraphics[width=\linewidth]{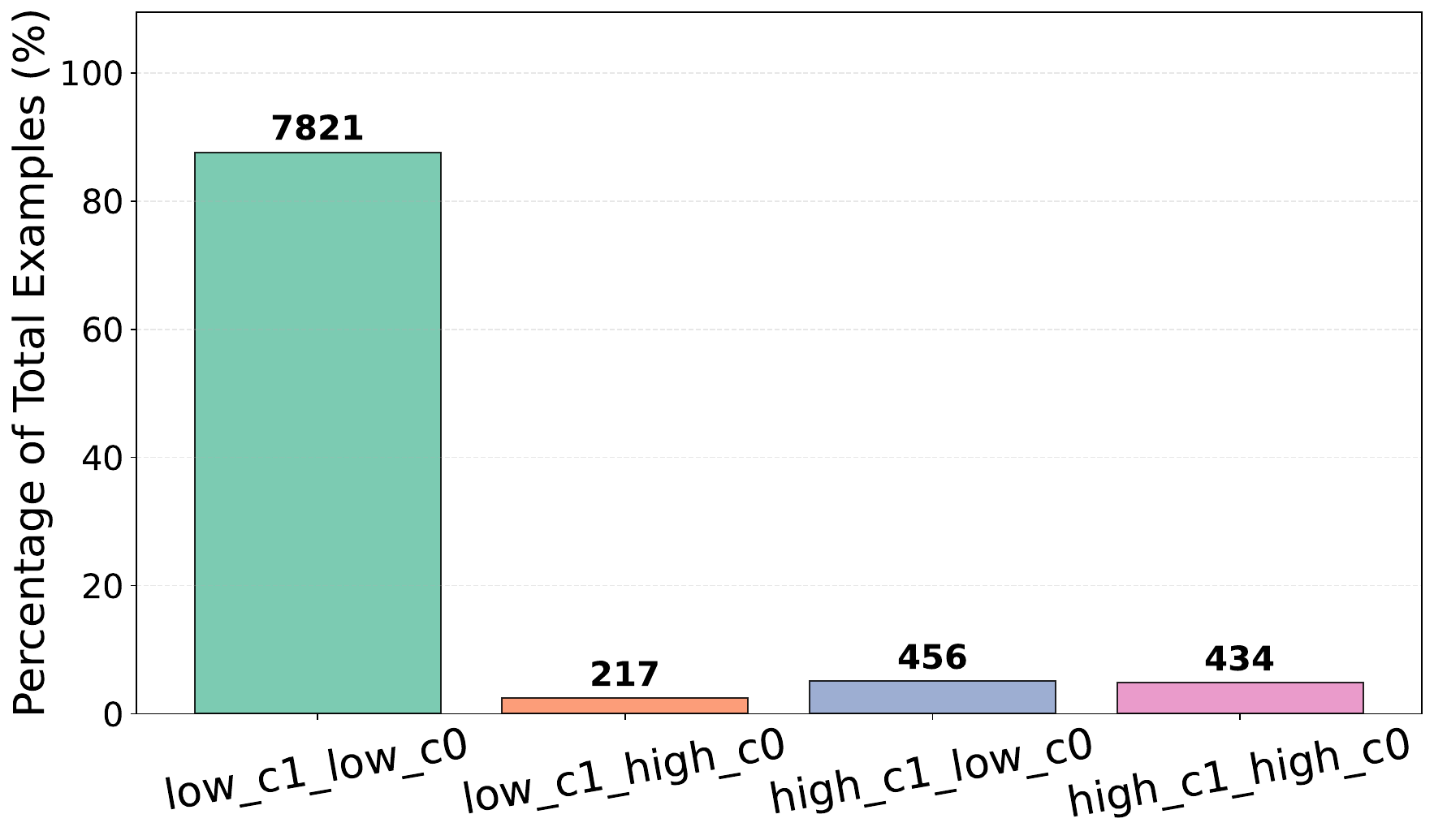}
    \caption{}
    \label{subfig:eval-ret-faith-b}
  \end{subfigure}
  \begin{subfigure}[t]{0.32\textwidth}
    \centering
    \includegraphics[width=\linewidth]{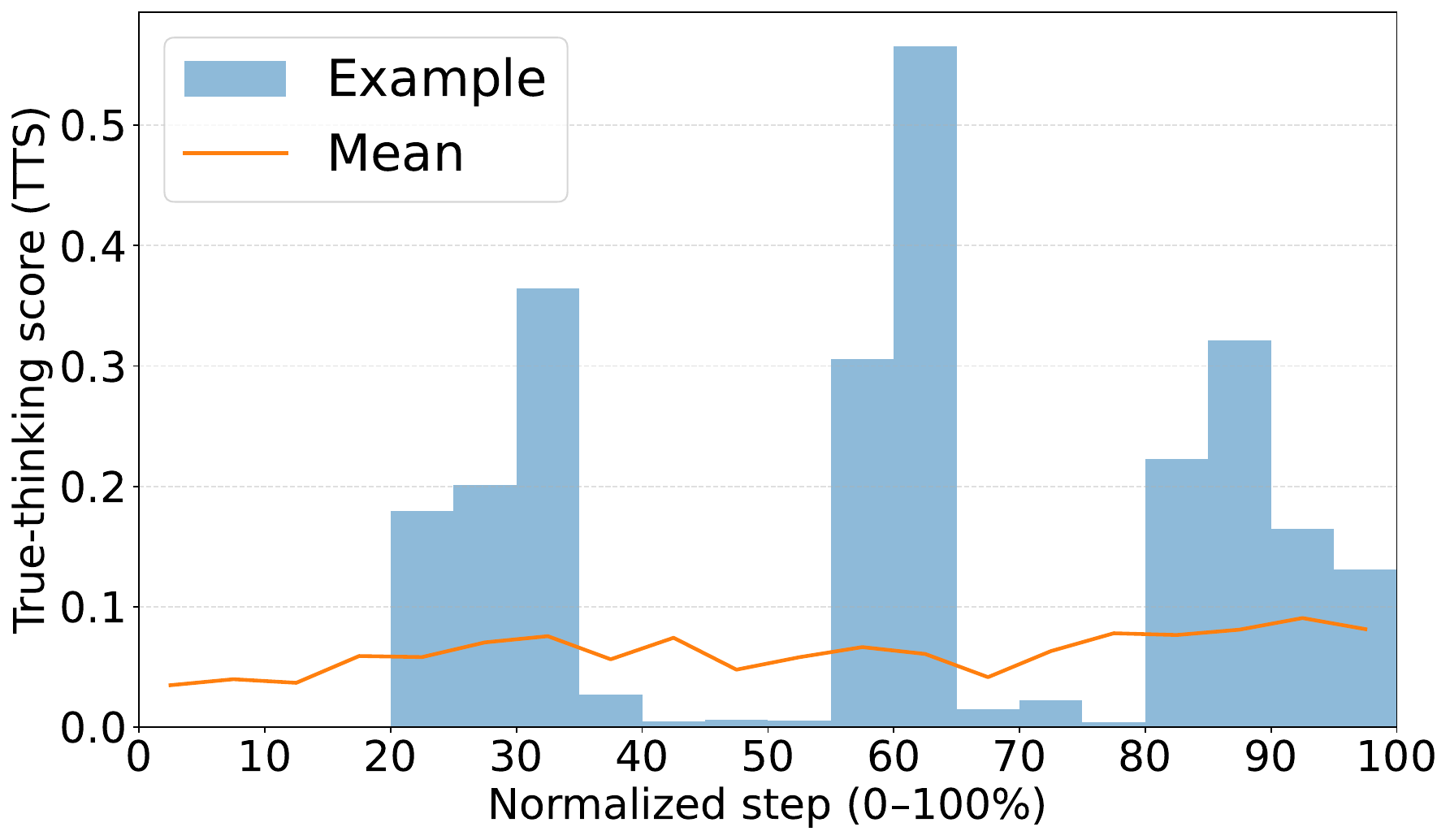}
    \caption{}
    \label{subfig:eval-ret-faith-c}
  \end{subfigure}
  \caption{ (a) The dataset-level distribution of the \scorename{} score; (b) The distribution for \ateNec{} and \ateSuf{} where \emph{low} means ATE($\cdot$) is below mean and \emph{high} means ATE($\cdot$) is above mean; (c) An example of \scorename{} for steps at different percentile and the respective mean \scorename{}. }\vspace{-5mm}
  \label{fig:eval-ret-faith}
\end{figure*}

\subsection{Context-based Average Treatment Effect}\label{sec:score}
The Average Treatment Effect (ATE)~\citep{rubin1974,pearl2009causality} quantifies the causal effect of a binary intervention $\mathbf{X}\in\{0,1\}$ on an outcome $\mathbf{Y}$ via Pearl's $\mathrm{do}(\cdot)$ operator:
\begin{align}
\mathrm{ATE} \;=\; \mathrm{S}_1 - \mathrm{S}_0,
\label{eq:ate}
\end{align}
where $\mathrm{S}_{x\in\{0,1\}} \;:=\; \Pr\!\big(\mathbf{Y}{=}1 \,\big|\, \mathrm{do}(\mathbf{X}{=}x)\big)$. To evaluate the causal contribution of a reasoning step $s$ for LLMs' reasoning, we condition on a \emph{context setting} $\mathbf{C}\in\{0,1\}$ and define a \emph{context-based ATE}:
\begin{align}
\mathrm{ATE}(c) &\;=\; \mathrm{S}_1(c) - \mathrm{S}_0(c), \\
\mathrm{S}_{x\in\{0,1\}}(c) &\;:=\; \Pr\!\big(\mathbf{Y}{=}1 \,\big|\, \mathbf{C}{=}c,\, \mathrm{do}(\mathbf{X}{=}x)\big),
\label{eq:cate}
\end{align}
where $c\in\{0,1\}$ specifies the context regime and $\mathbf{X}$ toggles using the intact step $s$ versus its perturbed counterpart $s'$. We consider two regimes:
a) \textit{Intact context} ($\mathbf{C}{=}1$): the original prefix $\mathcal{C}$ preceding $s$ is kept as generated;
b) \textit{Perturbed context} ($\mathbf{C}{=}0$): all steps in $\mathcal{C}$ are minimally perturbed (e.g., by small numeric offsets), weakening associations between $s$ and other steps so that the effect of $s$ can be isolated.

\textbf{Scoring the outcome.} Let $y^* := f(q,\mathcal{C}^{\star})$ be the model’s reference answer obtained via \emph{early-exit} on the full CoT, and let $f(q,\mathcal{C},s)$ denote the early-exit prediction after a given $(\mathcal{C},s)$. 
Instead of the binary $\mathbf{Y}=\mathbf{1}\{f(\cdot)=y^*\}$, we use the model’s confidence for the event $y^*$: $\Pr(\mathbf{Y}{=}1 \mid \cdot)\;\equiv\; \Pr\big(y^* \mid q,\mathcal{C},s\big)$, giving $\mathrm{S}_x(c)$ a probabilistic (confidence-based) interpretation.

\paragraph{Interpreting $\mathrm{ATE}_\text{nec}(1)$ vs.\ $\mathrm{ATE}_\text{suf}(0)$.} Conditioning on $\mathbf{C}$ allows us to distinguish two complementary notions of causal relevance:

\begin{itemize}[leftmargin=1.2em,topsep=2pt,itemsep=2pt]
    \item \textbf{Necessity under intact context ($\mathrm{ATE}_\text{nec}(1)$).}
    \begin{multline*}
\mathrm{ATE}_\text{nec}(1) = \Pr(y^* \mid \mathbf{C}{=}1,\,\mathrm{do}(\mathbf{X}{=}1)) \\
- \Pr(y^* \mid \mathbf{C}{=}1,\,\mathrm{do}(\mathbf{X}{=}0)).
\end{multline*}
\looseness=-1
    This tests whether $s$ is \emph{needed} given the full, supportive context $\mathcal{C}$. 
    A low $\mathrm{ATE}_\text{nec}(1)$ indicates that removing $s$ does not harm performance when other steps remain intact.

    \item \textbf{Sufficiency under perturbed context ($\mathrm{ATE}_\text{suf}(0)$).}
  \begin{multline*}
    \mathrm{ATE}_\text{suf}(0) 
    = \Pr(y^* \mid \mathbf{C}{=}0,\,\mathrm{do}(\mathbf{X}{=}1)) \\
    - \Pr(y^* \mid \mathbf{C}{=}0,\,\mathrm{do}(\mathbf{X}{=}0)).
 \end{multline*}
    This asks whether $s$ can \emph{on its own}---i.e., with weakened support from $\mathcal{C}$---drive the model toward $y^*$. 
    A high $\mathrm{ATE}_\text{suf}(0)$ suggests $s$ is \textbf{sufficiently informative} to elicit the correct answer, capturing causal relevance even when $s$ is not strictly necessary under the intact context.
\end{itemize}

\looseness=-100
Together, $\mathrm{ATE}_\text{nec}(1)$ (necessity) and $\mathrm{ATE}_\text{suf}(0)$ (sufficiency) provide a balanced view: A step can be causally meaningful by being necessary, sufficient, or both. 

\paragraph{True-Thinking Score (TTS).} Finally, the score is defined as 
\begin{equation}
\scorename(s)
= \tfrac{1}{2}\bigl(|\mathrm{ATE}_\text{nec}(1)| + | \mathrm{ATE}_\text{suf}(0)|\bigr). 
\label{eq:tts}
\end{equation}
A smaller \faithscore{s} indicates that the step has little causal influence on the model's prediction: perturbing or keeping it leads to almost the same result. Thus, that step is more likely to be \emph{decorative} rather than \emph{true thinking}. For each context setting \(c\), we measure the unsigned $\mathrm{ATE}(c)$,
\(|\mathrm{ATE}(c)| = |S_1(c)-S_0(c)|\). The sign of \(\mathrm{ATE}(c)\) reflects whether the step is helpful or harmful (e.g., the step is actually wrong) overall, but we are interested in \emph{how much} the model truly thinks through the step in its internal computation, regardless of direction. Taking the absolute value thus captures the magnitude of a step's causal effect and provides a broader measure of its importance. \scorename{} is a continuous metric, where lower values indicate that a step is more decorative.

\vspace{-2mm}
\section{Experimental Setup}
\vspace{-2mm}
\paragraph{Models.} We conduct experiments on many different families of open-source reasoning models that have strong reasoning abilities and can generate long CoTs. See details in Appendix~\ref{apdx:model}.


\vspace{-3mm}
\looseness=-1000
\paragraph{Data.} We evaluate on different reasoning benchmarks: {AMC} (American Mathematics Competitions), {AIME} (American Invitational Mathematics Examination) from 2020--2024, the {MATH} dataset~\citep{hendrycks2021measuring}, GSM8K~\citep{cobbe2021training}. In Section~\ref{sec:exp-steer}, we randomly sample 30\% of cases as a heldout test set, 10 \% of cases as validation set, and 60 \% of cases as a training set.

\section{Evaluation of Step-wise Causality}\label{sec:eval}

\looseness=-1000
As shown in Figure~\ref{fig:decorative-rate}, a significant fraction of CoT steps have \scorename{} below 0.005, suggesting that many steps have minimal causal impact on the model’s answer and may be decorative. Larger models generally achieve higher average \scorename{} and contain fewer decorative steps than smaller models, such as 4B or 27B models. However, scaling does not linearly improve \scorename{}: Kimi-K2.6 still has about 35\% decorative steps, while Nemotron3-Super-120B achieves the highest \scorename{}. These results suggest that a large fraction of CoT in current LLMs on practical reasoning questions may not be trusted as a faithful computation trace.


\vspace{-3mm}
\looseness=-1
\paragraph{The distribution of \scorename{} is long-tailed.}
As shown in Figure~\ref{subfig:eval-ret-faith-a} for Deepseek-R1-Distill-Qwen-2.5-7B on AIME, we find most steps have low scores, while only a few have very high scores. For example, as shown in Figure~\ref{subfig:eval-ret-faith-a} on the AIME dataset of Qwen-2.5, the mean is around 0.03. Only 6.4\% of CoT steps achieve a \scorename{} greater than 0.3, and merely 2.3\% exceed 0.7.  In Figure~\ref{subfig:eval-ret-faith-b}, we observe disparities between \ateNec{} and \ateSuf{} for the same step. This indicates that we need to consider both sufficiency and necessity to locate true-thinking steps. We confirm the same distribution pattern across different datasets (see Appendix~\ref{apdx:eval-faith-scale-model} for mathematical reasoning and results on Commonsense reasoning in Appendix~\ref{apdx:qa}).

\paragraph{Reasoning steps with high and low TTS are interleaved in a CoT.} Figure~\ref{subfig:eval-ret-faith-c} illustrates that steps with high \scorename{} scores can appear at different positions, though later steps are on average more likely to be true-thinking with higher \scorename{}. These results raise concerns about the reliability of monitoring LLMs by inspecting CoT~\citep{baker2025monitoring}, since individual steps may not always reflect the model’s true computation and contribute to the model's final answer.  



To further analyze low-\scorename{} steps, we categorize them using the rules in Appendix~\ref{sec:low_tts_taxonomy}. As shown in Figure~\ref{fig:low-tts-types}, different types of reasoning steps can be decorative, with intermediate calculations accounting for most cases. Notably, self-verification steps can also be decorative (example shown in Figure~\ref{fig:sufficiency-example-nemotron-nano}). Overall, \scorename{} quantifies the causal impact of reasoning traces and thus helps locate steps that may be redundant or may not reflect the model’s true reasoning process.

\subsection{Pruning CoT with \scorename{}}\label{sec:prune-cot}

\begin{figure}[ht]
    \centering
    \begin{subfigure}[t]{0.888\linewidth}
        \centering
        \includegraphics[width=\linewidth]{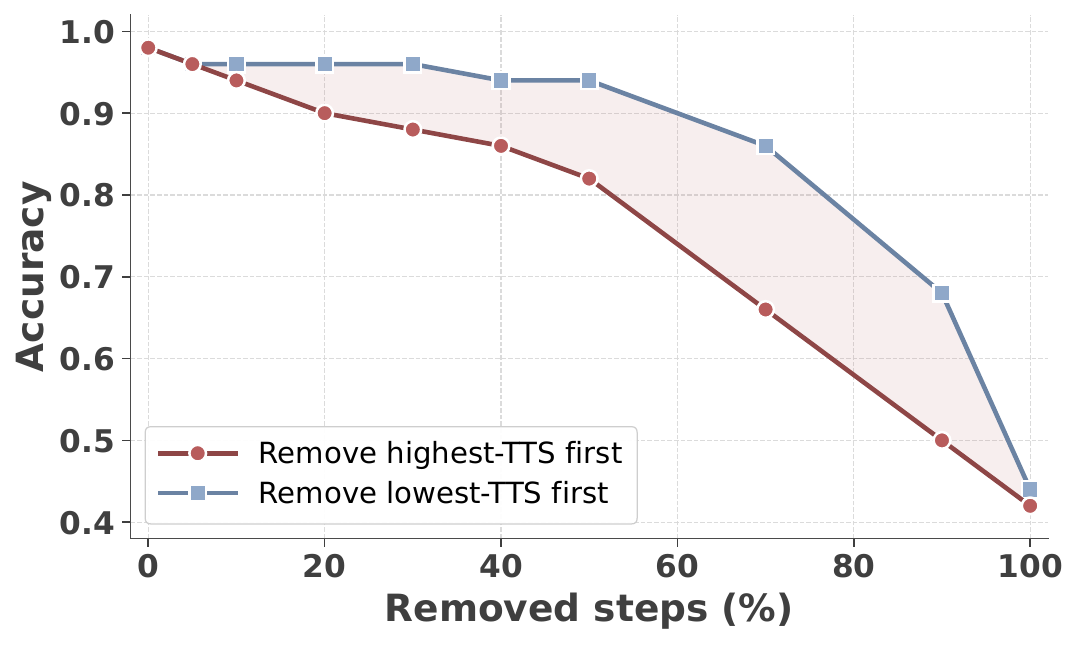}
        \caption{}
        \label{fig:tts-prune}
    \end{subfigure}
   
    \begin{subfigure}[t]{0.888\linewidth}
        \centering
        \includegraphics[width=\linewidth]{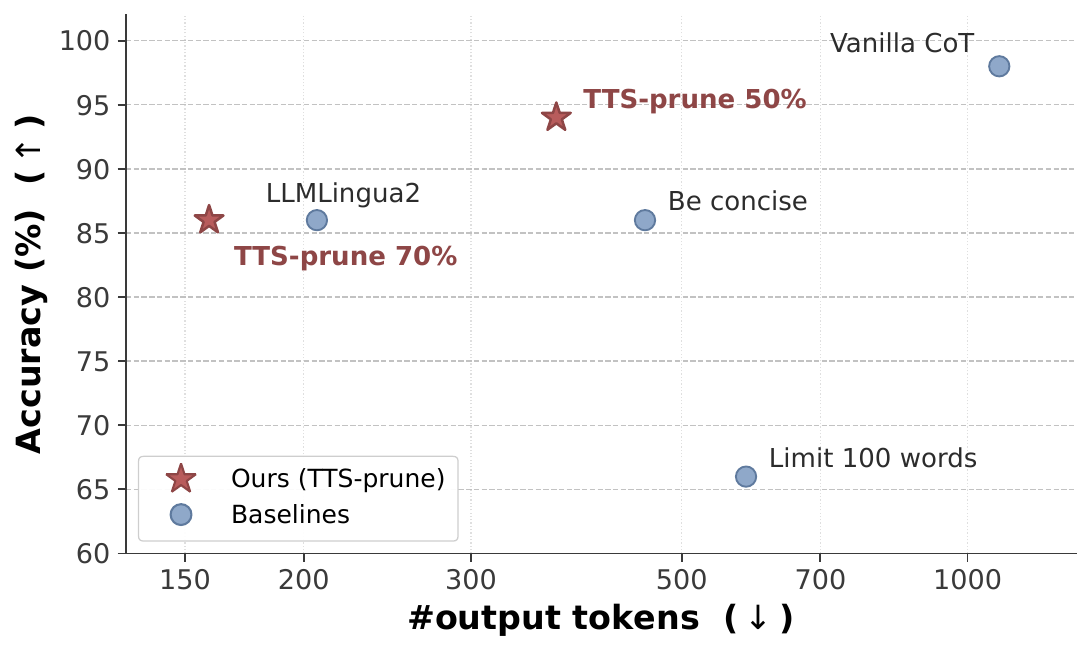}
        \caption{}
        \label{fig:pruned-cot-cmp}
    \end{subfigure}
    \caption{(a) Pruning reasoning steps in Nemotron3-Super-120B on MATH. Pruning the lowest TTS steps preserves accuracy far longer than pruning the highest TTS steps. (b) TTS-prune sits on the efficiency--accuracy frontier among different baselines.} \vspace{-5mm}
    \label{fig:tts-prompt-cmp}
\end{figure}

We further examine how pruning low- and high-\scorename{} CoT steps affects reasoning performance in this section.
\looseness=-1000
\paragraph{Pruning low-TTS steps retains the prompting performance.}  We rank all steps by \scorename{} in ascending order, remove the lowest or highest $r\%$ of steps, and reprompt the model with the pruned CoT. As shown in Figure~\ref{fig:tts-prune}, removing high-\scorename{} steps quickly degrades performance, whereas pruning up to half of the lowest-\scorename{} steps preserves most of the performance. Furthermore, we compare pruning the lowest-\scorename{} steps with other methods for producing shorter CoT: (1) \textit{Be concise}~\citep{renze2024benefits}, which directly prompts the model to be concise; (2) \textit{Limit the answer length to 100 words}~\citep{han2025token}, which prompts LLMs to control the generation budget. This can shorten CoT, though generations may still exceed the token limit; (3) LLMLingua-2, a trained LM that labels semantically important tokens to remove the unimportant ones~\citep{pan2024llmlingua,xia2025tokenskip}. As shown in Figure~\ref{fig:pruned-cot-cmp}, pruning 50\% of the steps from the lowest \scorename{} provides a good balance between token length and reasoning performance compared with these methods. Pruning 70\% further shortens the CoT, but sacrifices performance a bit. 

\begin{table}[ht]
\centering
\small
\setlength{\tabcolsep}{6pt}
\resizebox{\columnwidth}{!}{%
\begin{tabular}{l cc cc}
\toprule
& \multicolumn{2}{c}{\textbf{MATH}} & \multicolumn{2}{c}{\textbf{GSM8K}} \\
\cmidrule(lr){2-3} \cmidrule(lr){4-5}
\textbf{Variant} & Acc.\ $\uparrow$ & \#Token\ $\downarrow$ & Acc.\ $\uparrow$ & \#Token\ $\downarrow$ \\
\midrule
Base              & 88 & 732 & 76 & 349 \\ \hline
\begin{tabular}[c]{@{}l@{}}TokenSkip \\ \citep{xia2025tokenskip}\end{tabular}
& 84 & 686 & 76 & 308 \\

\begin{tabular}[c]{@{}l@{}}ST-ICL \\ \citep{munkhbat2025self}\end{tabular}
& 79 & 496 & 79 & 324 \\
ST-TTS-prune (ours)  & \textbf{88} & \textbf{488} & \textbf{87} & \textbf{279} \\
\bottomrule
\end{tabular}}
\caption{Performance of different training methods to improve reasoning efficiency. }\vspace{-5mm}
\label{tab:sft-pruning}
\end{table}

\looseness=-1000
\paragraph{Further self-training can improve reasoning efficiency.} We further demonstrate that self-training (ST)~\citep{xie2020self} on TTS-pruned CoT (pruning rate=50\%) can reduce the model's CoT length, while preserving the performance. We compare against TokenSkip~\citep{xia2025tokenskip}, which trains on CoT pruned by LLMLingua-2, and ST-ICL~\citep{munkhbat2025self}, which self-trains on concise CoT generated through in-context learning. We use LoRA~\citep{hu2022lora} (rank=32) to fine-tune Nemotron3-nano-30B on 300 pruned CoT examples from MATH for five epochs, and evaluate on held-out sampled test cases of MATH and GSM8K. 

As shown in Table~\ref{tab:sft-pruning}, ST-\scorename{}-prune produces the fewest reasoning tokens while maintaining the best reasoning performance among the compared methods. On MATH, it reduces reasoning tokens by 66\% relative to the base model, and on GSM8K it also improves performance. In contrast, TokenSkip only modestly reduces reasoning length, while ST-ICL substantially degrades performance. Neither baseline appears to generalize well to GSM8K in terms of improving reasoning efficiency.

Overall, \scorename{} provides a useful signal for pruning CoT that can be used in student-model training~\citep{guo2025deepseek} or self-training for more efficient reasoning~\citep{munkhbat2025self}. It may also be explored as a reward signal in reinforcement learning~\citep{guo2025deepseek,lambert2024tulu} to reduce steps with low causal impact and improve CoT efficiency and trustworthiness, which we leave to future work.

\begin{table*}[t]
\centering
\small
\setlength{\tabcolsep}{4.5pt}
\renewcommand{\arraystretch}{1.18}
\newcommand{\na}{\textemdash}
\begin{tabular}{ll *{2}{S} *{2}{S} *{2}{S} *{2}{S}}
\toprule
\multicolumn{2}{c}{\textbf{Dataset / Method}} &
\multicolumn{2}{c}{\textbf{Attention scaling}} &
\multicolumn{2}{c}{\textbf{Random vector}} &
\multicolumn{2}{c}{\textbf{DropStep direction}} &
\multicolumn{2}{c}{\textbf{\direction{} direction}} \\
\cmidrule(lr){1-2}\cmidrule(lr){3-4}\cmidrule(lr){5-6}\cmidrule(lr){7-8}\cmidrule(lr){9-10}
&& \multicolumn{1}{c}{ET} & \multicolumn{1}{c}{DT}
 & \multicolumn{1}{c}{ET} & \multicolumn{1}{c}{DT}
 & \multicolumn{1}{c}{ET} & \multicolumn{1}{c}{DT}
 & \multicolumn{1}{c}{ET} & \multicolumn{1}{c}{DT} \\
\midrule
\multicolumn{10}{l}{\emph{AMC}} \\
& Qwen-2.5
& 6.2 & 25.0 & 4.0 & 26.9 & 1.5 & 28.6 & \textbf{55.0} & \textbf{55.7} \\
& Llama-3.1
& \textbf{24.8} & 20.5 & 3.5 & 20.6 & 10.5 & 32.4 & 17.6 & \textbf{35.3} \\
& Nemotron
& 5.1 & 27.2 & 4.5 & 45.5 & 9.0 & 45.4 & \textbf{35.7} & \textbf{54.5} \\
\addlinespace
\multicolumn{10}{l}{\emph{MATH}} \\
& Qwen-2.5
& 10.0 & 23.9 & 2.0 & 30.2 & 2.5 & 17.7 & \textbf{49.8} & \textbf{69.2} \\
& Llama-3.1
& 7.5 & 35.4 & 5.0 & 47.9 & 11.0 & 52.1 & \textbf{14.0} & \textbf{54.2} \\
& Nemotron
&21.7 &42.7 & 21.5 &44.6 & 6.5 &45.1 & \textbf{59.5}  & \textbf{56.3}\\
\addlinespace
\multicolumn{10}{l}{\emph{AIME}} \\
& Qwen-2.5
& 9.3 & 25.0 & 1.5 & 21.4 & 1.5 & 14.3 & \textbf{55.5} & \textbf{53.6} \\
& Llama-3.1
& 6.3 & 35.2 & 2.5 & 29.4 & 5.0 & 41.1 & \textbf{38.0} & \textbf{47.1} \\
& Nemotron
& 12.0 & 70.6 & 6.5 & 76.5 & 4.5 & 79.5 & \textbf{39.0} & \textbf{91.2} \\
\bottomrule
\end{tabular}

\caption{Top-1 flip rate among all layers (\%) $\uparrow$ in the \testa{} (ET) and the \testb{} (DT). We use flip rate as the metric, measuring how often steering changes the model's initial prediction. AMC dataset is in-domain evaluation where \direction{} directions are extracted, while the other two datasets are for out-of-domain evaluation. } \vspace{-4mm}
\label{tab:intervention-results}
\end{table*}

\begin{figure*}[t]
  \centering
  \begin{subfigure}[t]{0.2\textwidth}
    \centering
    \includegraphics[width=\linewidth]{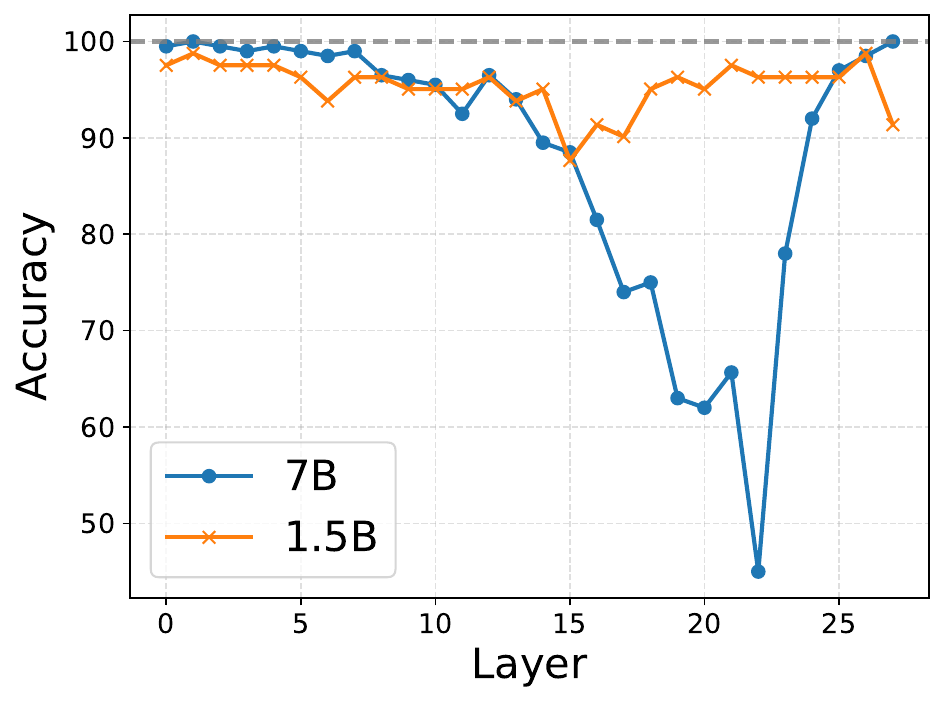}
    \caption{ \testa{} }
    \label{subfig:steer-a}
  \end{subfigure}
  \begin{subfigure}[t]{0.2\textwidth}
    \centering
    \includegraphics[width=\linewidth]{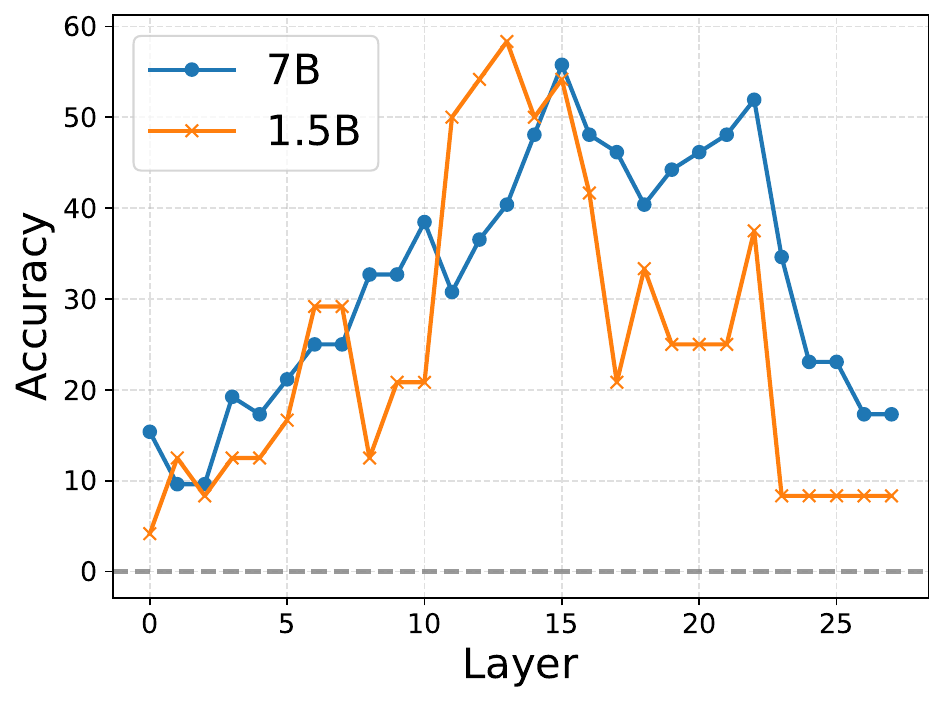}
    \caption{ \testb{}}
    \label{subfig:steer-b}
  \end{subfigure}
  \begin{subfigure}[t]{0.2\textwidth}
    \centering
    \includegraphics[width=\linewidth]{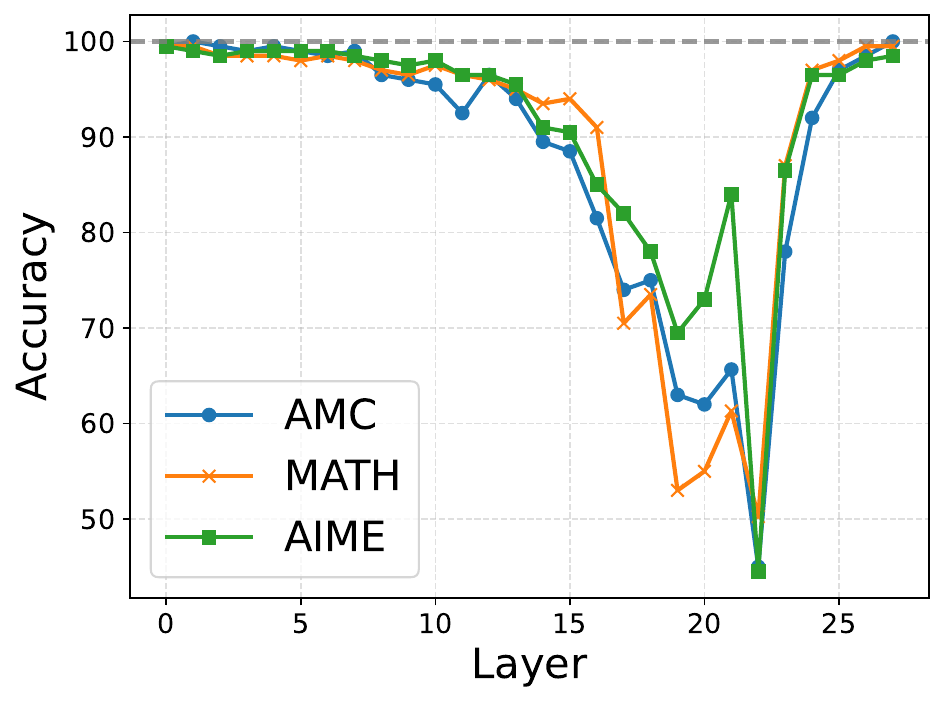}
    \caption{ \testa{}}
    \label{subfig:steer-c}
  \end{subfigure}
  \begin{subfigure}[t]{0.2\textwidth}
    \centering
    \includegraphics[width=\linewidth]{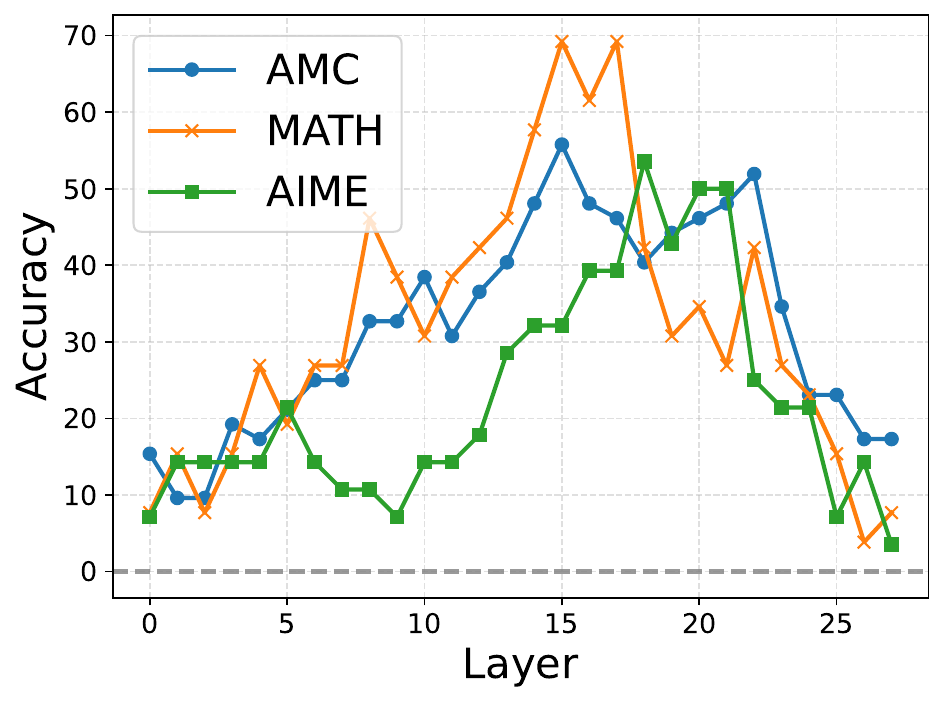}
    \caption{ \testb{}}
    \label{subfig:steer-d}
  \end{subfigure}
  \looseness=-100
  \caption{Layer-wise results of steering with the \direction{} vector. In the \testa{}, stronger intervention is reflected by lower accuracy (more right→wrong flips); In the \testb{}, by higher accuracy (more wrong→right flips). Figures (a–b): layer-wise results on AMC for DeepSeek-R1-Distill-Qwen-7B and its 1.5B variant under the \testa{} and the \testb{}. Figures (c–d): cross-domain results, where the \direction{} direction is extracted on AMC and applied to MATH and AIME. } \vspace{-3mm}
  \label{fig:steering}
\end{figure*}

\section{Mediating Reasoning via Steering}\label{sec:exp-steer}
In this section, we further provide mechanism insights on decorative thinking in CoT through steering. We empirically reveal that, in LLMs, latent-space steering with the \direction{} direction identified by \scorename{} can mediate whether to use a reasoning step in CoT when computing the results. Steering the hidden states of a step along \direction{} direction can force LLMs to engage with that step to compute its answer and may increase the attention to that step as well and vice versa.

\vspace{-2mm}
\paragraph{Extracting the steering direction with \scorename{}.}\label{sec:method-dir} 
We first detail how to extract the steering direction. We focus on the residual stream activation ${h}^{l}(s_t)$ of the last token position $t$ for a step $s$ at a layer $l$.  At a layer $l$, we collect the hidden states of the most representative steps $s_\text{TT}$ (where \faithscore{s_\text{TT}} $\geq$ threshold $\alpha$) and steps $s_\text{DT}$ (where \faithscore{s_\text{DT}} $\leq$ threshold $\beta$). Following the difference-in-means approach~\citep{marks2023geometry,arditi2024refusal,zhao2025llms}, we compute the direction as the mean shift from $\mu^{l}_{\text{TT}}=\text{mean}(h^{l}(s_{\text{TT}}))$ to $\mu^{l}_{\text{DT}}=\text{mean}(h^{l}(s_{\text{DT}}))$ in the latent space, i.e., $v^l_{\direction{}} = \mu^l_{\mathrm{TT}} - \mu^l_{\mathrm{DT}}.$ For steering at test time,  we modify the residual stream for the hidden state of a test step in the example by using activation addition at a single layer $l$, i.e., ${\bar{h}}^{l}={h}^{l}+v^{l}_{\direction{}}$ to all tokens in the step. Detailed implementations are in Appendix~\ref{apdx:steering-imp}.

Specifically, we have two steering cases, \textbf{(1) Engagement Test}: We consider cases where $f(q,\mathcal{C})=y_{GT}$ and $f(q,\mathcal{C},s')=y_{GT}$. Namely, the model still obtains the ground-truth answer $y_{GT}$ despite the perturbed step $s'$.  If we apply a steering direction $v_{}^l$ to the hidden state of $s'$, and the model's correct answer flips to an incorrect one ($f^{+v_{}^l}(q,\mathcal{C},s')\neq y_{GT}$), this indicates that the steering has forced the model to reason over $s'$, thus following the errors injected into $s'$. \textbf{(2) Disengagement Test}: On the other hand, we consider the reverse cases where the model first predicts the correct answer, i.e., $f(q,\mathcal{C})=y_{GT}$, but including a perturbed step $s'$ causes it to fail: $f(q,\mathcal{C},s')\neq y_{GT}$. If applying a steering direction $-v_{}^l$ to $s'$ flips the wrong answer to the correct answer ($f^{-v_{}^l}(q,\mathcal{C},s')=y_{GT}$), then the intervention has made the model disregard the step $s'$.

\looseness=-100
As comparison baselines, we consider three approaches for layer-wise intervention apart from the \direction{} direction. (1) DropStep direction: Motivated by the prior work~\citep{tanneru2024hardness,lanham2023measuring,bogdan2025thought,wang-etal-2025-chain} on faithfulness evaluation, we compare prediction confidence before and after dropping step $s$, i.e., $\mathrm{P}(y_{GT}|q,\mathcal{C},o)-\mathrm{P}(y_{GT}|q,\mathcal{C})$ (equivalent to $\mathrm{ATE}_\text{nec}(1)$ in our framework). The steps of the highest and lowest scores are then used to extract a steering direction following the same method as the \direction{} direction.  (2) Random steering vector: we generate a random vector with the same dimensionality and norm as the \direction{} direction to test whether our identified direction encodes meaningful information. (3) Attention scaling: we scale the attention weights of the tokens of step $s$ at a layer during inference time, with scale=100 encouraging the model to think through the step and scale=0 suppressing it. 

\vspace{-2mm}
\subsection{Whether to use a reasoning step can be steerable}\label{sec:steering-ret}

\looseness=-1000
As shown in Table~\ref{tab:intervention-results}, steering with the (reverse) \direction{} direction reliably flips predictions in both tests. In the \testb{}, it effectively prevents the model from using the perturbed step $s'$, with effects far stronger than those of random vectors. This shows that suppression of step use with the reverse \direction{} direction in the \testb{} arises from a meaningful signal rather than noise. Additionally, the \direction{} direction derived from our \scorename{} can produce stronger and more consistent intervention effects than the steering directions extracted using DropStep. 

On the other hand, our experiments across datasets show that {the latent signal controlling whether a step engages in reasoning is universal}. As seen in Table~\ref{tab:intervention-results}, the \direction{} direction extracted on AMC generalizes well to other datasets across all models, indicating a model-internal mechanism of thinking rather than a dataset-specific artifact. In the Qwen-2.5 model, layer 15-22 consistently yield the strongest intervention performance across all three datasets (Figure~\ref{subfig:steer-c}–\ref{subfig:steer-d}), suggesting these intermediate layers may concentrate latent reasoning. We also compare models with different sizes in the same model family. Similar patterns hold for the much smaller Qwen-2.5-1.5B model (Figure~\ref{subfig:steer-a} and Figure~\ref{subfig:steer-b}) and larger 14B model (Figure~\ref{fig:steer-14b} in Appendix~\ref{apdx:14b}). Steering along the \direction{} direction (to induce the step use in LLMs' internal reasoning) is weaker for smaller models, whereas the results in the \testb{} are comparable.

\looseness=-1000
\paragraph{Steering with the \direction{} direction mediates LLMs' attention.} We find that steering with the \direction{} direction can reallocate attention among tokens. In \testb{}, steering in the reverse \direction{} direction reduces attention to the perturbed tokens (see examples in Figures~\ref{fig:attn-map-c0} and~\ref{fig:attn-map-d0} of the Appendix), causing the model to largely disregard them. Similarly, steering along the \direction{} direction increases attention to the step (see examples in Figure~\ref{fig:attn-map-a} and Figure~\ref{fig:attn-map-b} of the Appendix). Overall, when steering along the \direction{} direction, the attention of the steered tokens will increase on average by 193\% and steering along the reverse direction will decrease the attention by 32.3\% across the layers. 


\paragraph{Steering LLMs for true self-verification.}\label{sec:steer-self-check}
We define decorative-thinking self-verification $\mathcal{S}_\text{DT-self-verify}=(s_1, s_2,\dots,s_n)$ where $\text{TTS}(s_i)$ is smaller than 0.005. We focus on a setting in which the model fails to follow correct steps in self-verification $\mathcal{S}_\text{DT-self-verify}$ and is instead misled by incorrect previous steps $\mathcal{C}'$, resulting in a wrong prediction, i.e., $f(q,\mathcal{C}',\mathcal{S}_\text{DT-self-verify}) \neq y_{GT}$. We apply the \direction{} direction to the tokens in $\mathcal{S}_\text{DT-self-verify}$. We find that this can at best recover around $51\%$ of the cases where the models do not follow their correct self-verification steps in CoT and give wrong answers (layer-wise results shown in Figure~\ref{fig:steer-unfaith-self-verify} of Appendix). Remarkably, the layer with the strongest steering effect aligns with the layer identified in Figure~\ref{subfig:steer-c}–\ref{subfig:steer-d}, suggesting that certain layers may play a special role in determining whether reasoning steps are engaged in the model's internal computation.

\section{Related Work} 
\looseness=-1000
\paragraph{Internal thinking in LLMs' latent space.} Apart from explicit CoT, LLMs ``think'' internally across their layers. They can directly answer reasoning problems, sometimes even matching the performance of CoT-based prompting~\cite{ma2025reasoning}. Prior work suggests that LLMs may solve certain tasks through internal \emph{circuits}~\cite{yang2024large,marks2024sparse,prakash2025language}. The \emph{implicit} reasoning capabilities of LLMs may bypass using explicit CoTs~\citep{deng2023implicitcot,hao2024coconut,pfau2024let,goyal2023think}. Such internal reasoning of LLMs questions how much the model truly uses each step in their CoT, which we study using a causal framework.

\looseness=-1000
\paragraph{The unfaithfulness of LLMs' reasoning traces.} Prior work has raised broad concerns about the faithfulness of reasoning traces. Models may rely on hidden hints or spurious biases without acknowledging them in CoT~\citep{chen2025reasoning,arcuschin2025chain,chua2025deepseek,turpin2023language}, and generated rationales may be insensitive to perturbations or unnecessary for reaching the final answer~\citep{gaoshapley,lanham2023measuring,tanneru2024hardness,wang-etal-2025-chain}. While these studies provide important evidence that CoT can be unfaithful or post-hoc, they mostly evaluate CoT holistically, leaving the causal role of individual steps underexplored.

\looseness=-1000
\paragraph{Steering LLMs.} Steering directions applied to intermediate hidden states have been found to mediate LLMs' behaviors or perception in many aspects~\citep{von2024language,turner2023steering,tigges2023linear,li2023inference,marks2023geometry,zhao2025llms}.  In terms of reasoning, past works have found steering vectors that can control the strength of reasoning, e.g., longer or shorter CoT~\citep{tang2025glore,sun2025thinkedit,chen2025seal,sheng2025reasoningstrength} or different reasoning styles in CoT~\citep{venhoff2025understanding}. 

\section{Conclusions}

\looseness=-1000
We propose True-Thinking Score to analyze step-wise causality in CoT and show that frontier LLMs contain many decorative steps that make minimal causal contribution to the model's answer on common reasoning benchmarks. We reveal that \emph{true-thinking} and \emph{decorative-thinking} steps are often interleaved: only a small subset causally influences the final prediction, while many steps merely create the appearance of reasoning. We further show that \scorename{} can prune CoTs into concise rationales for self-training, substantially improving reasoning efficiency without sacrificing performance. Mechanistically, we demonstrate that whether a reasoning step contributes to the model's computation can be mediated by a \direction{} direction, suggesting that step-level use is reflected in internal representations. Broadly, our findings raise concerns about both the efficiency of CoT reasoning of current LLMs and the reliability of using CoT to interpret LLMs' internal thought.


\section{Limitations} 
Our causal evaluation framework is inherently approximate and especially focuses on math reasoning that has strong logic. It is greedy in nature and may not capture all possible causal pathways, nor does it aim to reconstruct a complete causal graph of reasoning steps. Thus, it should be viewed as a probe that highlights representative \emph{true-thinking} and \emph{decorative-thinking} steps rather than a definitive oracle of internal reasoning. However, we can still leverage \scorename{} to prune CoT effectively. In addition, the \direction{} direction we extract may not be optimal. We regard our findings as an existence proof that whether to engage with a reasoning step can be mediated by steering directions, and we leave the development of more effective directions and a deeper understanding of their geometry to future work.  



\section{Acknowledgment}
We appreciate the support from the MATS program in hosting the research project and providing computing resources.

\bibliography{ref}

\clearpage
\newpage
\appendix

\begin{figure}[!t]
\centering
\footnotesize
\begin{tcolorbox}[colback=gray!5, colframe=gray!40, boxrule=0.5pt, arc=2pt, left=4pt, right=4pt, top=3pt, bottom=3pt]

\textbf{Problem.} Simplify $\dfrac{\sec x}{\sin x} - \dfrac{\sin x}{\cos x}$.

\tcblower

\textbf{Preceding context:}
\begin{quote}\scriptsize\ttfamily
...Express everything in terms of $\sin$ and $\cos$. Common denominator: $\sin x / \cos x = \sin^2 x / (\sin x \cos x)$. Difference: $(1 - \sin^2 x)/(\sin x \cos x) = \cos^2 x/(\sin x \cos x) = \cos x / \sin x = \cot x$.
\end{quote}

\textbf{Original step:}
\begin{quote}\scriptsize\ttfamily
Check: $\sec x / \sin x = 1/(\cos x \sin x)$ $\sin x / \cos x = \sin^2 x / (\cos x \sin x)$ $(1 - \sin^2 x)/(\cos x \sin x) = \cos^2 x/(\cos x \sin x)$
\end{quote}

\textbf{Perturbed step:}
\begin{quote}\scriptsize\ttfamily
Check: $\sec x / \sin x = 4/(\cos x \sin x)$ $\sin x / \cos x = (\sin^{-1} x)/(\cos x \sin x)$ $(-2 - \sin^5 x)/(\cos x \sin x) = \cos^1 x/(\cos x \sin x)$
\end{quote}
\end{tcolorbox}
\caption{ Example decorative verification step of Nemotron3-Nano. Perturbing the step with intact and perturbed context shifts confidence of model's predicted answer ($\cot x$) by only $6.4\text{e-}12$ and $4.5\text{e-}14$ respectively.}
\label{fig:sufficiency-example-nemotron-nano}
\end{figure}

\begin{figure}[ht]
  \centering
  \includegraphics[width=0.6\linewidth]{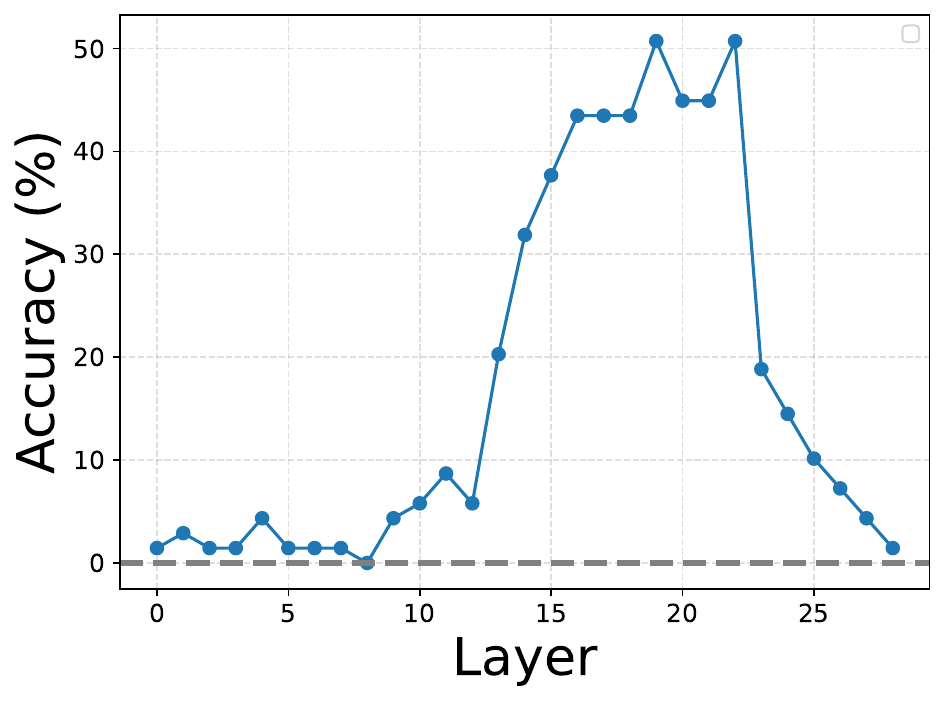}
  \captionsetup{skip=4pt} 
  \caption{Layer-wise results after steering the model to truly self-verify their reasoning steps. Initially, the test accuracy is zero without steering, as the model ignores its self-verification steps in CoT and outputs wrong answers.}\vspace{-6mm}
  \label{fig:steer-unfaith-self-verify}
\end{figure}

\section{Taxonomy of Low-TTS Reasoning Steps}
  \label{sec:low_tts_taxonomy}

To better understand which reasoning steps receive low \scorename{}, we categorize low-TTS steps into a small set of functional types. This taxonomy is intended as an interpretability tool rather than a learned classifier. Each step is assigned to exactly one category using a priority-ordered rule set based on lexical and structural cues, so that broad classes such as calculation do not absorb more specific functions such as verification or restatement.

\paragraph{Verification.}
Steps that explicitly check or validate a previous result, with cues such as ``check,'' ``verify,'' ``substitute,'' or ``sanity check.'' 

\paragraph{Restatement.}
Steps that restate an intermediate result/ condition or summarize a derivation, often with cues such as ``therefore,'' ``thus''. 

\paragraph{Calculation.}
Steps that contain explicit numerical or algebraic manipulation, identified by mathematical operators, equations, fractions, or digits. This class is particularly important: low-TTS calculations show that even symbolic-looking computation can be causally inert. In many cases, perturbing numbers in these steps leaves the final answer unchanged, suggesting that the calculation is redundant or post-hoc.

\paragraph{Setup.}
Steps that introduce notation, variables, or representations, with cues such as ``let,'' ``define,'' ``denote,'' ``set,'' or ``represent.'' Low-TTS setup steps often correspond to conventional problem translation: they may make the solution easier to read, but the model may not rely on them to compute the answer.

\paragraph{Alternative route.}
Steps that introduce an optional solution path, with cues such as ``alternative,'' ``another way,'' or ``could also.'' 

\paragraph{Other.}
Steps that do not match the above patterns are assigned to an ``other'' category. This residual bucket captures heterogeneous cases and is not interpreted as a coherent reasoning type.



\clearpage
\newpage

\begin{figure*}[t]
  \centering
  \begin{subfigure}[t]{0.4\textwidth}
    \centering
    \includegraphics[width=\linewidth]{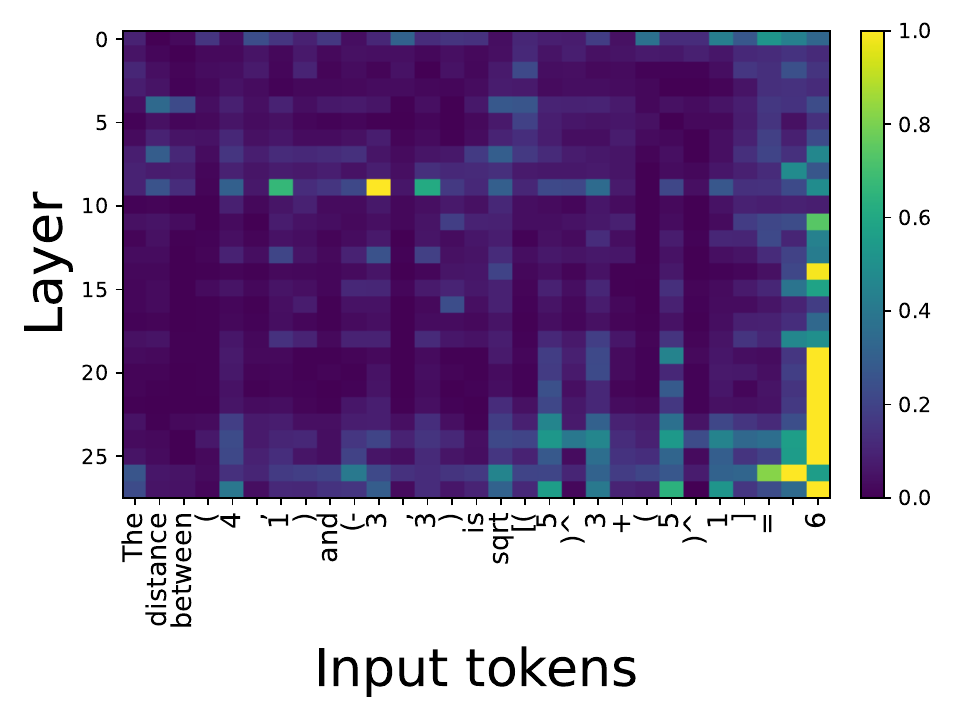}
    \caption{ \testa{}: After steering}
    \label{fig:attn-map-a}
  \end{subfigure}
  \begin{subfigure}[t]{0.4\textwidth}
    \centering
    \includegraphics[width=\linewidth]{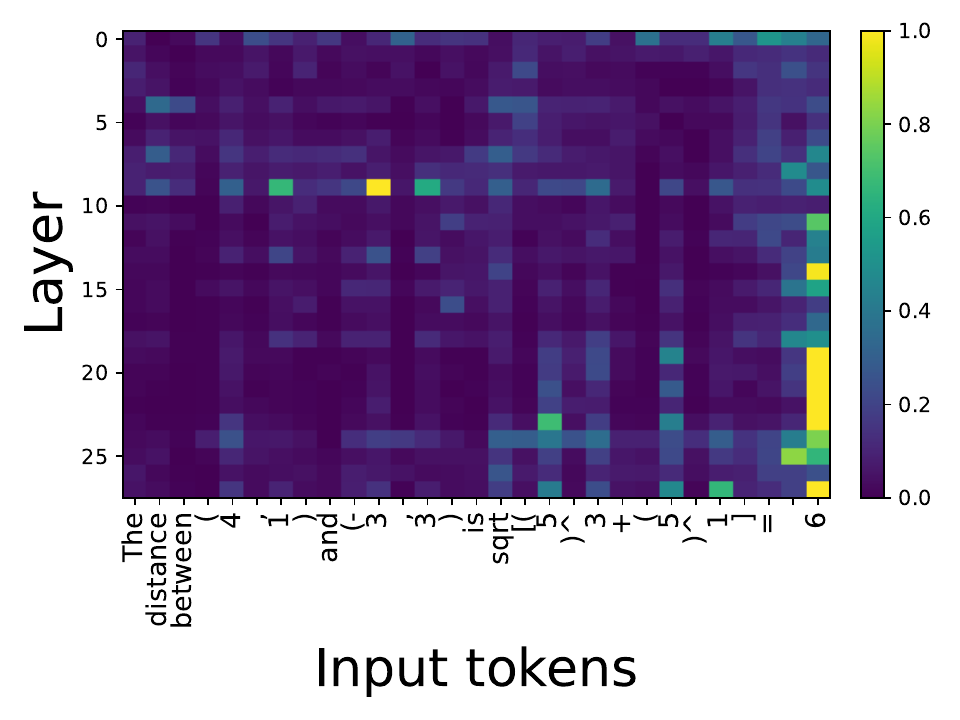}
    \caption{ \testa{}: Before steering}
    \label{fig:attn-map-b}
  \end{subfigure}

    \begin{subfigure}[t]{0.4\textwidth}
    \centering
    \includegraphics[width=\linewidth]{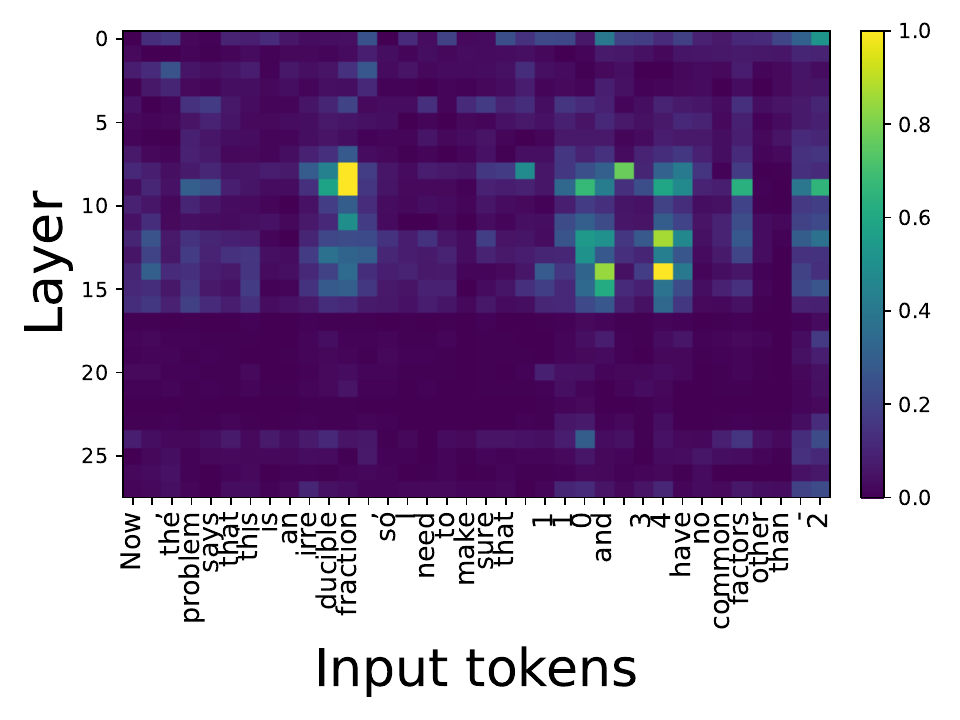}
    \caption{ \testb{}: After steering}
    \label{fig:attn-map-c0}
  \end{subfigure}
  \begin{subfigure}[t]{0.4\textwidth}
    \centering
    \includegraphics[width=\linewidth]{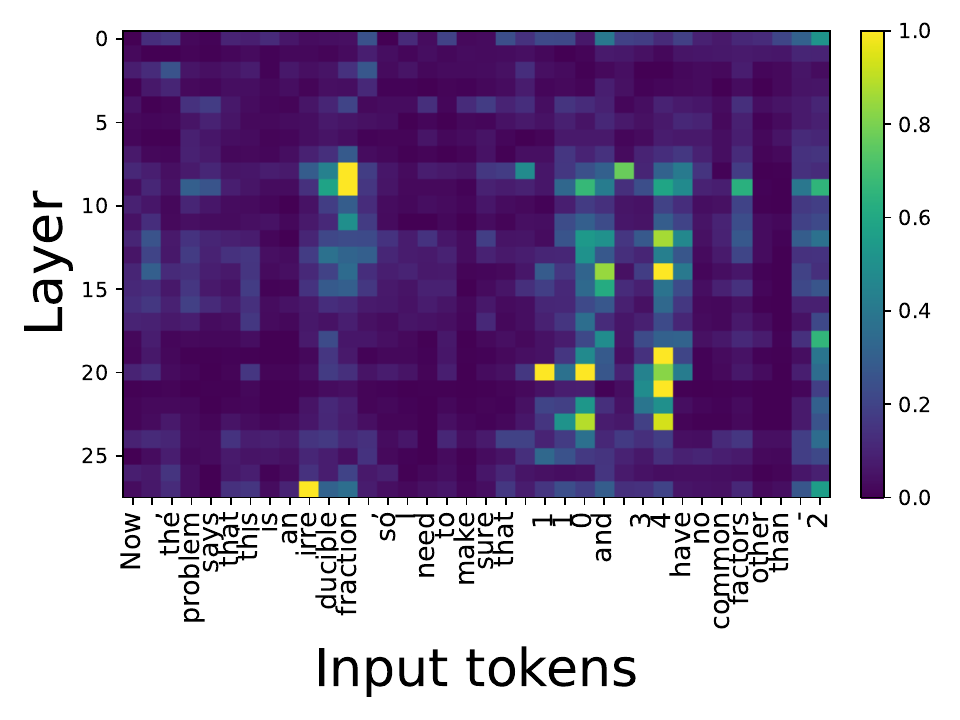}
    \caption{ \testb{}: Before steering}
    \label{fig:attn-map-d0}
  \end{subfigure}
  \caption{Normalized attention scores of the step in the \testa{} and the \testb{} before and after steering at Layer 17. (a–b) Applying the \direction{} direction to a step increases the model’s attention to it. (c–d) Applying the reverse \direction{} direction decreases the model’s attention.}
  \label{fig:attn-map-apdx}
\end{figure*}

\section{More related work}\label{apdx:relate-work}

\paragraph{CoT pruning.} Our work is also related to CoT compression~\citep{xia2025tokenskip,song2025reasoning,kang2025c3ot,yu2025causal} and important steps analysis~\citep{golovneva2022roscoe}. However, those works are different as they focus on the semantics of verbalized steps for analysis and do not look into the underlying causality. For example, \citet{xia2025tokenskip} prune long CoTs into shorter sentences with equivalent meanings for LLMs without degrading performance. But that does not mean pruned steps do not causally contribute to the model's answer. Our work may suggest similar results that there exist shorter alternative CoTs leading the model to get the same answers, but they do not solve whether a step truly engages with the model's computation, and the internal mechanism of LLMs' thinking process is still unclear. \citet{yu2025causal} reframe CoT pruning into the Probability of Necessity and Sufficiency (PNS) framework~\citep{pearl2009causality}, but it focuses on estimating the necessity of each step using counterfactual intervention in a similar way to prior works~\citep{lanham2023measuring,yee2024dissociation,bogdan2025thought}. Although they are framed into PNS, their measurement of sufficiency and necessity is restricted to contexts where the prior steps remain unchanged, which corresponds to $\mathrm{ATE}_\text{nec}(1)$ in our framework.


\section{Implementations}
\label{apdx:imp}

\paragraph{Perturbing reasoning steps.} We treat sentences as distinct reasoning steps, as prior work has shown that each sentence can serve a different function within a reasoning trace~\citep{bogdan2025thought}. We follow prior work~\citep{lanham2023measuring,gaoshapley} and add small random offsets (chosen from $[-3,-2,-1,1,2,3]$) to the numbers in a reasoning step. This keeps the perturbation minimal so that the step remains largely unchanged in token length, wording, and underlying logic. We can therefore reasonably attribute any confidence changes caused by the perturbation to the model’s treatment of the original step. For steps that do not contain numerical values, we also follow prior work~\citep{bogdan2025thought,tanneru2024hardness,wang-etal-2025-chain} by dropping them as a perturbation to measure the influence of those sentences. For perturbing context steps, we only change numerical values.

\subsection{Models}\label{apdx:model}

We evaluate on different open-weight LLMs spanning 1.5B to 1T parameters. For Qwen-2.5-7B and Llama-3.1-8B, we use the versions finetuned on samples generated by DeepSeek-R1~\citep{guo2025deepseek}, i.e., DeepSeek-R1-Distill-Qwen-7B and DeepSeek-R1-Distill-Llama-8B, along with Nemotron-1.5B; these models generate very long CoTs at test time. From the latest Qwen3 family~\citep{yang2025qwen3}, we include Qwen3.5-4B, Qwen3.5-27B, and Qwen3.6-35B-A3B. We additionally evaluate Nemotron3-Nano-30B and Nemotron3-Super-120B~\citep{blakeman2025nvidia}, GPT-OSS-120B~\citep{agarwal2025gpt}, DeepSeek-V3.1-671B~\citep{liu2024deepseek}, and Kimi-K2.6-1T\footnote{\url{https://huggingface.co/moonshotai/Kimi-K2.6}}. This set covers dense and Mixture-of-Experts architectures, transformer and hybrid Mamba-Transformer designs, and both reasoning-tuned and general-purpose post-training recipes. 

\subsection{Steering}\label{apdx:steering-imp}
To compute the \direction{} direction, for all tested models we set the threshold $\alpha$=0.9 for $s_\text{TT}$, while $\beta$=0 for $s_\text{DT}$, which means perturbing those steps $s_\text{DT}$ does not change the model's confidence at all. We emphasize that thresholding is used only to select the most representative steps for extracting the steering direction, while \scorename{} is a continuous measure. The ablation study for thresholding is in Appendix~\ref{apdx:threshold-ablation}.

\section{More Experimental Results of \scorename{} }\label{apdx:eval-faith-scale-model}

\begin{figure*}[t]
  \centering
  \begin{subfigure}[t]{0.4\textwidth}
    \centering
    \includegraphics[width=\linewidth]{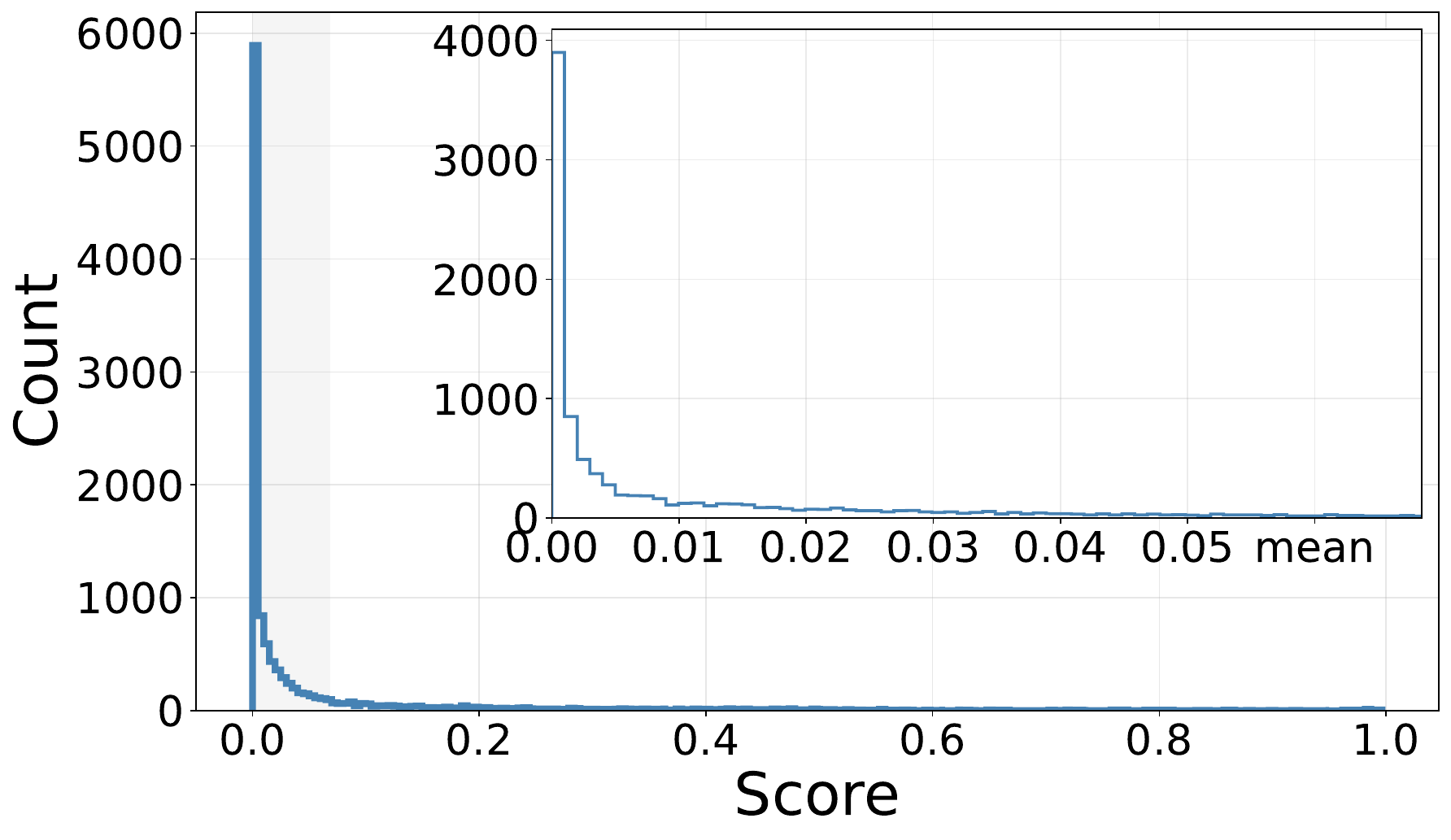}
    \caption{MATH dataset}
    \label{fig:eval-tts-apdx-subfig1}
  \end{subfigure}
  \begin{subfigure}[t]{0.4\textwidth}
    \centering
    \includegraphics[width=\linewidth]{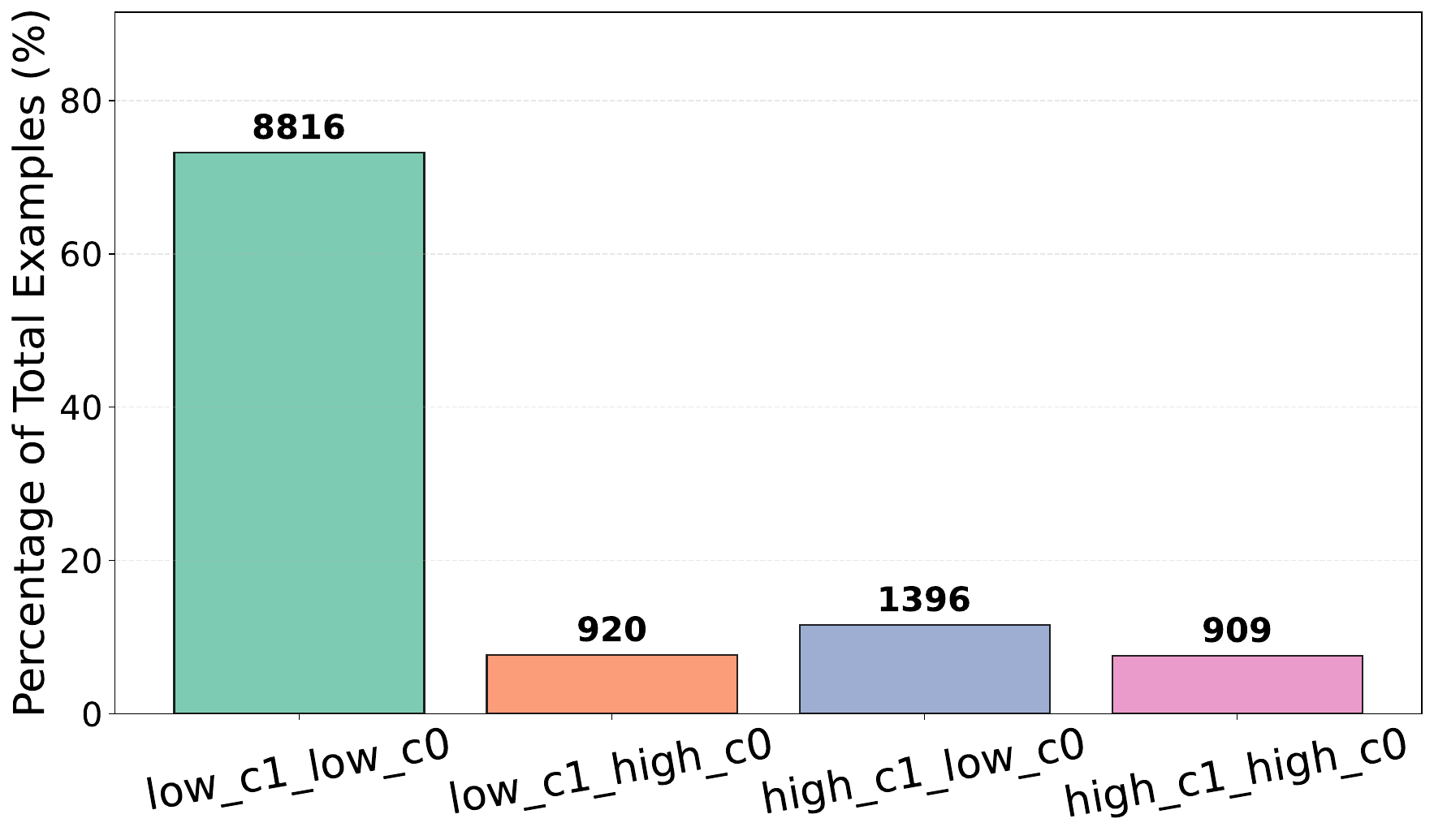}
    \caption{MATH dataset}
    \label{fig:eval-tts-apdx-subfig2}
  \end{subfigure}
  
  \begin{subfigure}[t]{0.4\textwidth}
    \centering
    \includegraphics[width=\linewidth]{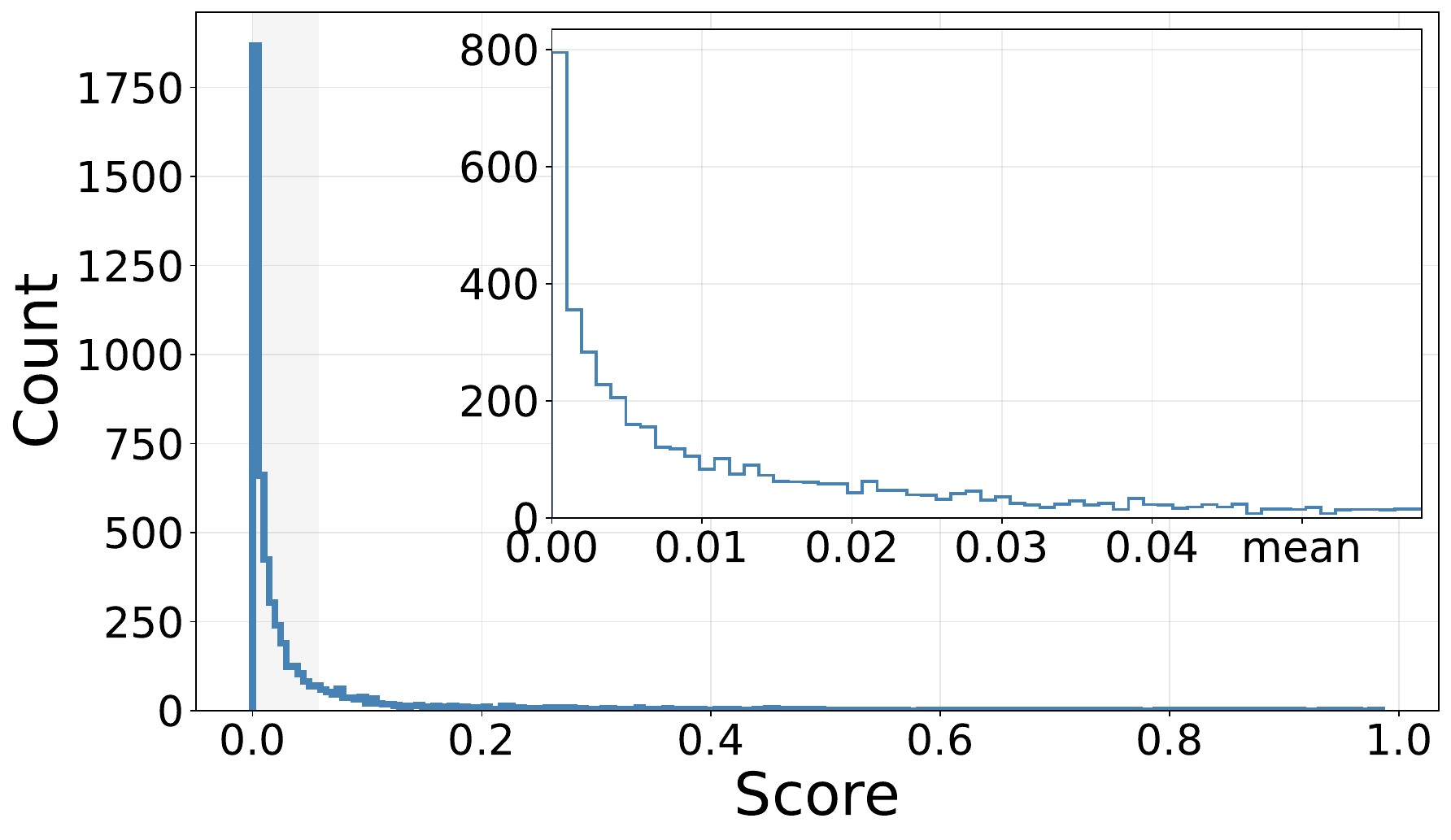}
    \caption{AMC dataset}
    \label{fig:eval-tts-apdx-subfig3}
  \end{subfigure}
  \begin{subfigure}[t]{0.4\textwidth}
    \centering
    \includegraphics[width=\linewidth]{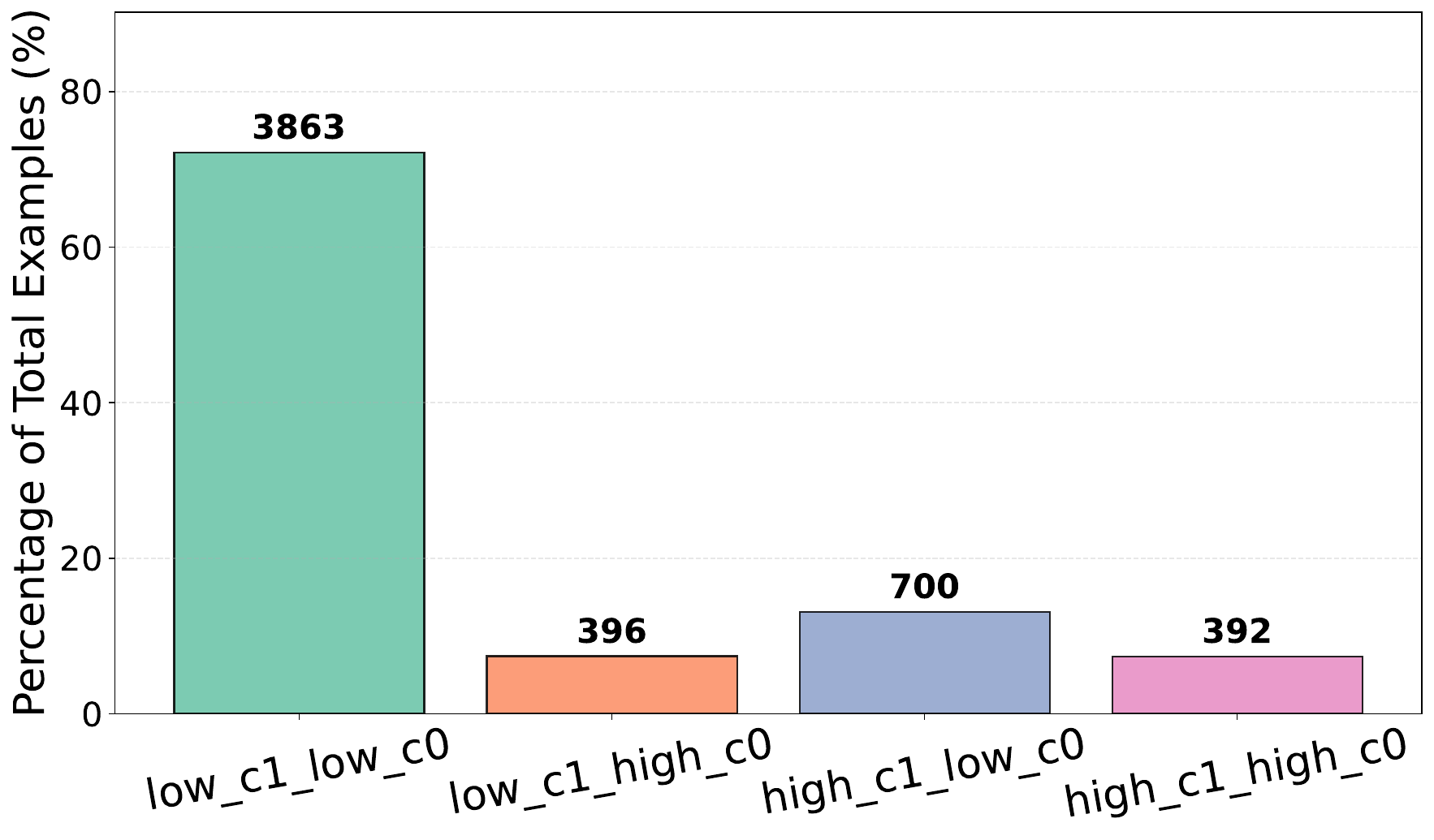}
    \caption{AMC dataset}
    \label{fig:eval-tts-apdx-subfig4}
  \end{subfigure}
  
  \caption{Distribution of \scorename{} on different datasets. }
  \label{fig:eval-tts-apdx-qwen}
\end{figure*}

As shown in Figure~\ref{fig:eval-tts-apdx-qwen}, we observe a similar long-tail distribution of  \scorename{} across different datasets. Steps with larger \scorename{} are sparse, while the most of steps have relatively low \scorename{}.  Additionally, as shown in Figure~\ref{fig:eval-tts-apdx-subfig2} and Figure~\ref{fig:eval-tts-apdx-subfig4}, on different datasets, we still observe that there exist ample examples where $\mathrm{ATE}_\text{nec}(1)$ and $\mathrm{ATE}_\text{suf}(0)$ of the same step are very different.

\begin{figure*}[t]
  \centering
  \begin{subfigure}[t]{0.32\textwidth}
    \centering
    \includegraphics[width=\linewidth]{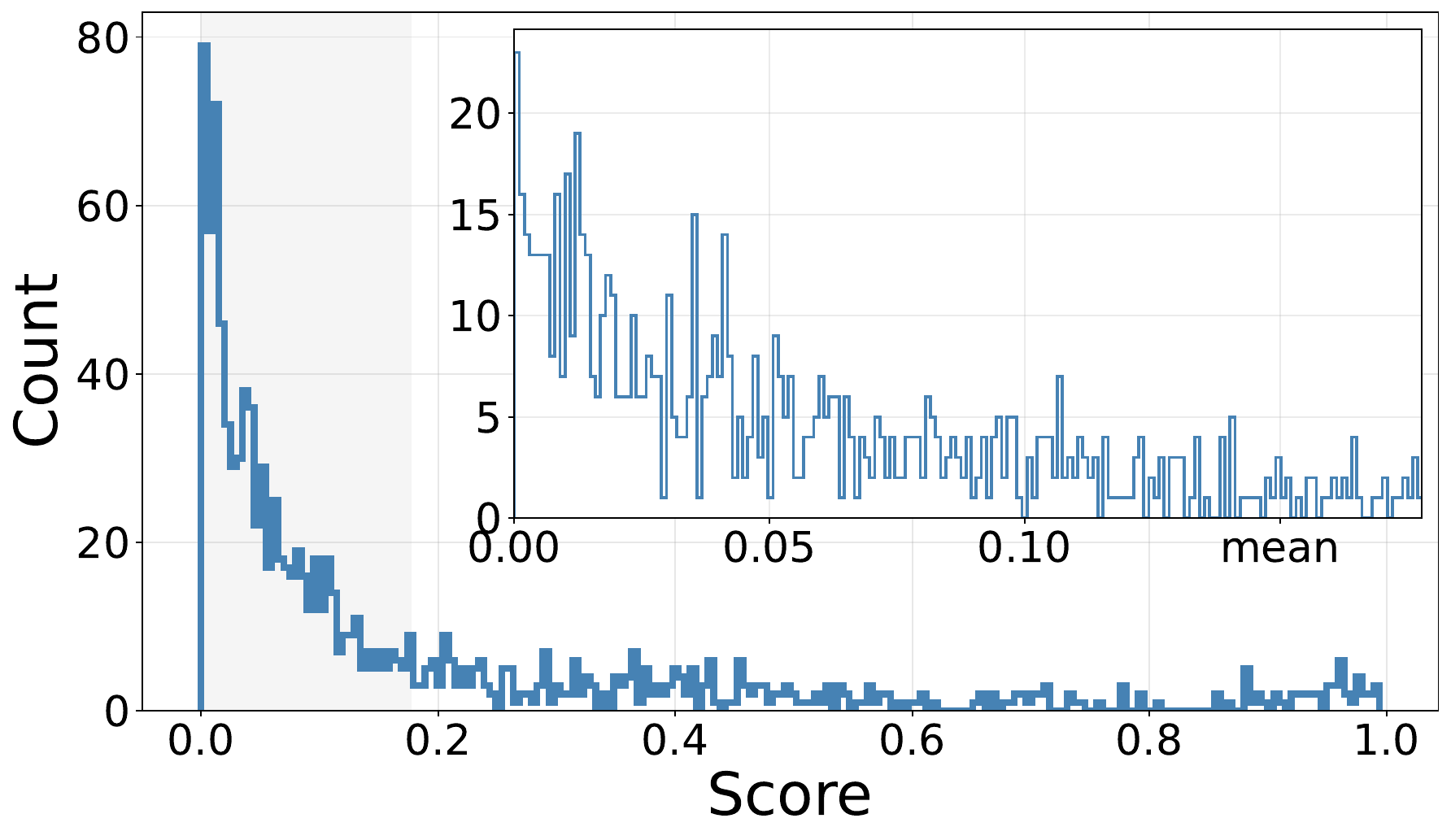}
    \caption{}
    \label{subfig:eval-ret-qa-a}
  \end{subfigure}
  \begin{subfigure}[t]{0.32\textwidth}
    \centering
    \includegraphics[width=\linewidth]{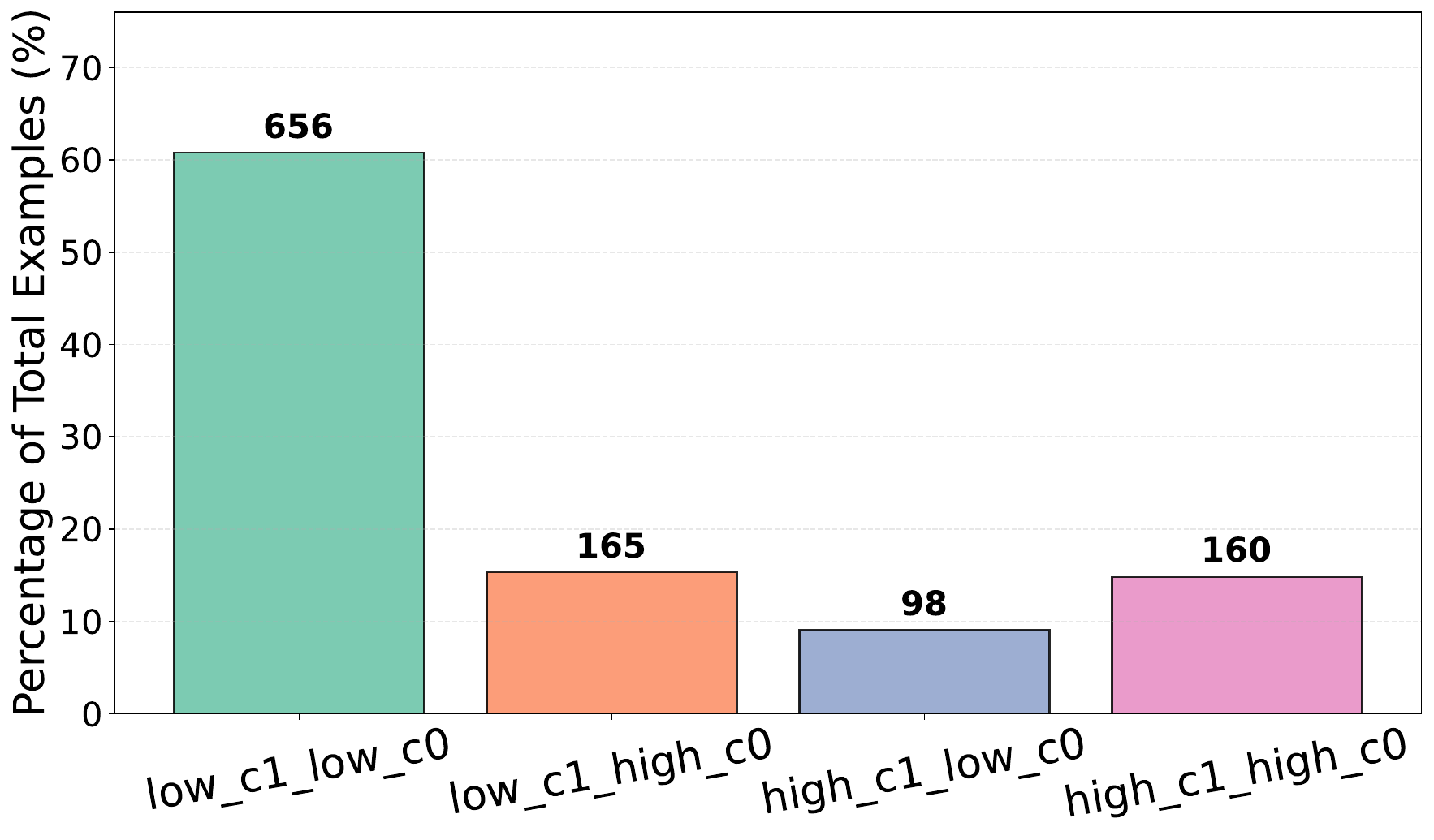}
    \caption{}
    \label{subfig:eval-ret-qa-b}
  \end{subfigure}
  \begin{subfigure}[t]{0.32\textwidth}
    \centering
    \includegraphics[width=\linewidth]{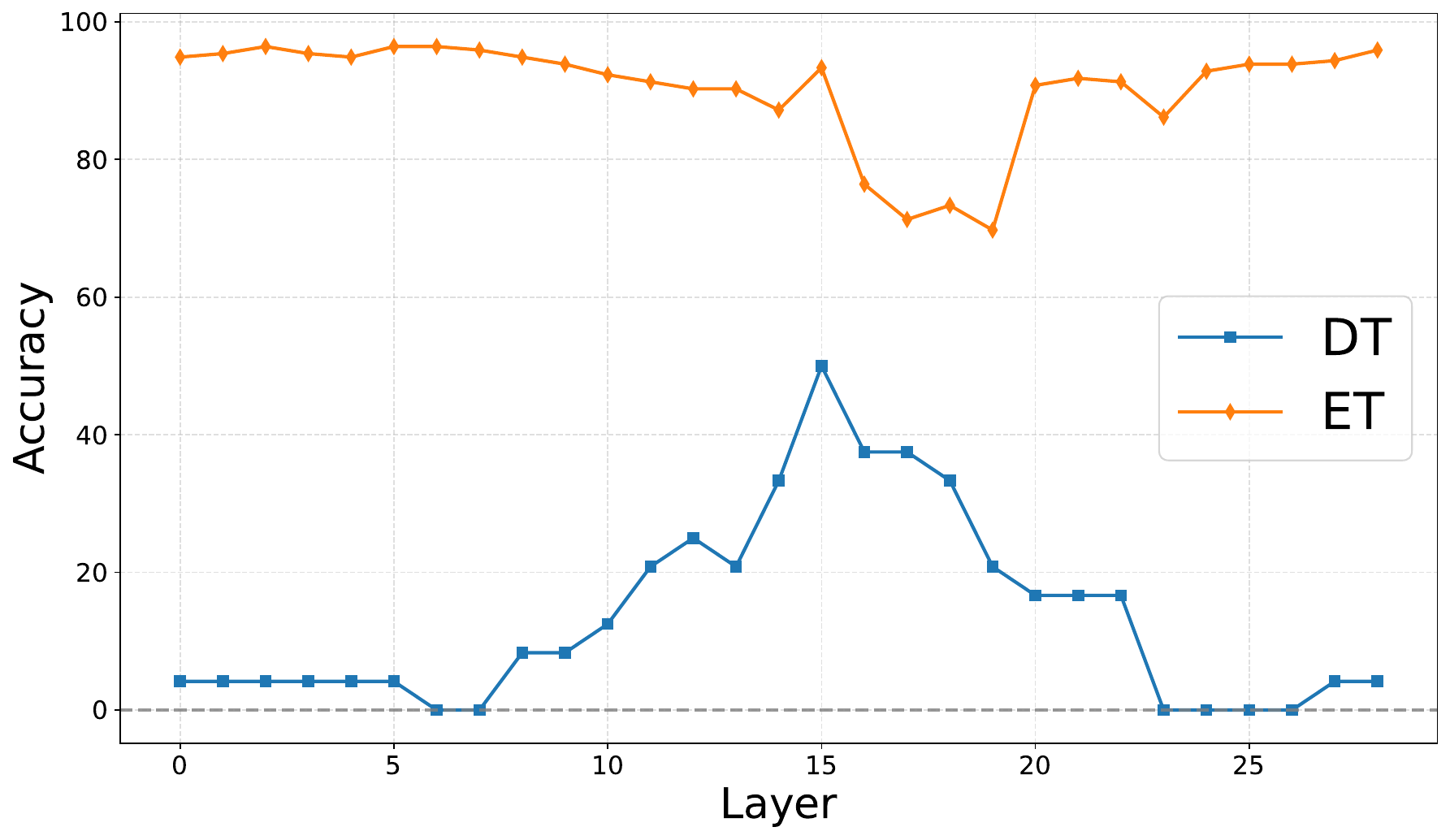}
    \caption{}
    \label{subfig:eval-ret-qa-c}
  \end{subfigure}
  \caption{ Results for Commonsense reasoning. (a) The dataset-level distribution of the \scorename{} score; (b) The distribution for $\mathrm{ATE}_\text{nec}(1)$ and $\mathrm{ATE}_\text{suf}(0)$ where \emph{low} means ATE($\cdot$) is below mean and \emph{high} means ATE($\cdot$) is above mean; (c) Steering results on the Engagement Test and the Disengagement Test. }\vspace{-3mm}
  \label{fig:commonqa}
\end{figure*}

\section{Results on Commonsense Reasoning}\label{apdx:qa}

We apply \scorename{} and the \direction{} direction to commonsense reasoning to test the generalizability of our framework to more generic reasoning cases.  We use CommonsenseQA~\citep{talmor2019commonsenseqa} as our test data. Since commonsense reasoning is predominantly non-numeric, standard intervention schemes that rely on perturbing numerical values are not suitable in this setting. To apply interventions on such content, we prompt a separate large language model (GPT-5) to produce subtle, semantically coherent distortions of the initial reasoning step.  This makes perturbation applicable to different reasoning scenarios. An example perturbed rationale is shown in Table~\ref{tab:rationale-example-qa}.

\begin{table}[t]
\centering
\small
\setlength{\tabcolsep}{6pt}
\begin{tabularx}{\linewidth}{@{}lX@{}}
\toprule
\textbf{Question} &
Where would you find magazines alongside many other printed works? \\[2pt]
\textbf{Options} &
\textbf{A} doctor \qquad
\textbf{B} \checkmark~{bookstore} \qquad
\textbf{C} market \qquad
\textbf{D} train station \qquad
\textbf{E} mortuary \\[6pt]
\textbf{Original rationale} &
...\underline{bookstores} are designed to sell both books and periodicals... \\[4pt]
\textbf{Perturbed rationale} &
...\underline{train stations} are designed to sell both books and periodicals... \\
\bottomrule
\end{tabularx}
\caption{Example multiple-choice question with original vs.\ perturbed rationale.}
\label{tab:rationale-example-qa}
\end{table}

As shown in Figure~\ref{subfig:eval-ret-qa-a} and Figure~\ref{subfig:eval-ret-qa-b}, the distribution of \scorename{} on Commonsense reasoning is long-tailed, which is similar to the patterns observed on mathematical reasoning. 

Furthermore, we conduct the steering tests (explained in Section~\ref{sec:exp-steer}) on the examples of Commonsense QA. Specifically, we use the same \direction{} direction extracted from AMC dataset for steering (the same setting as Table~\ref{tab:intervention-results}). The steering results are shown in Figure~\ref{subfig:eval-ret-qa-c}. Noticeably, the \direction{} direction can still mediate the reasoning behaviors on CommonsenseQA when applied to the middle layers. This strongly indicates the generalizability of \direction{} direction and suggests that LLMs encode a specific variable for true-thinking behaviors.

\begin{figure}
    \centering
    \includegraphics[width=0.8\linewidth]{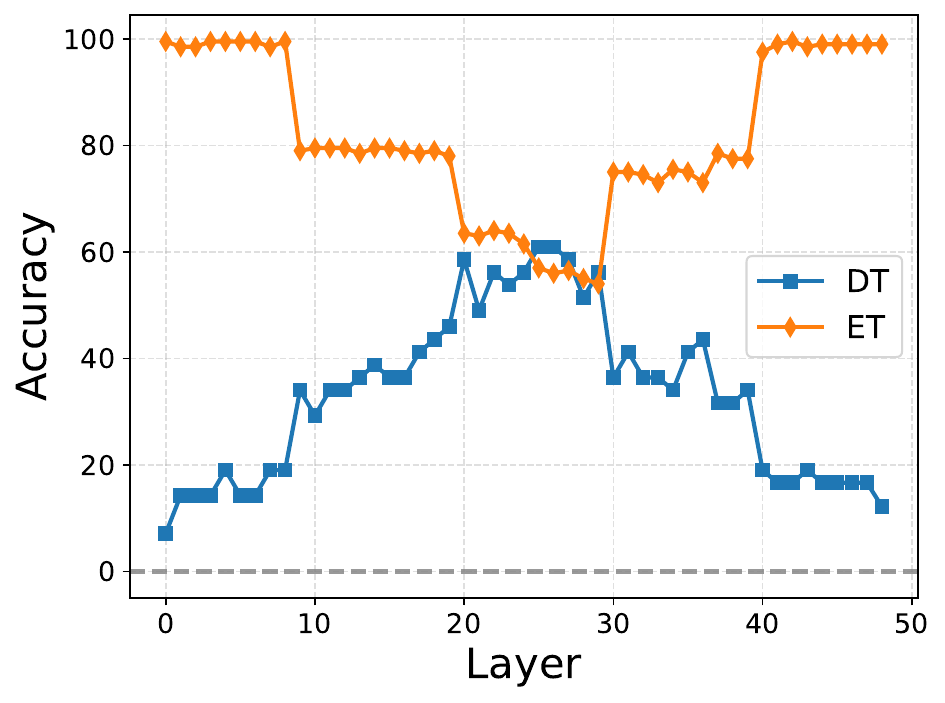}
    \caption{Results of \testa{} and \testb{} on Qwen-14B.}
    \label{fig:steer-14b}
\end{figure}
\section{Steering Results on Larger Models}\label{apdx:14b}
Beyond the 1.5B and 7B models in Figure~\ref{subfig:steer-a} and Figure~\ref{subfig:steer-b}, we further demonstrate that the \direction{} direction generalizes to a larger 14B model. As shown in Figure~\ref{fig:steer-14b}, we can similarly modulate the 14B model’s reasoning behavior, controlling whether it engages or disengages with reasoning steps by steering the hidden states in the middle layers along or against \direction{} direction.

For instance, in the Qwen-2.5 model, layer 15-22 consistently yield the strongest intervention performance across all three datasets (Figure~\ref{subfig:steer-c}–\ref{subfig:steer-d}), suggesting these intermediate layers concentrate latent reasoning. We also compare models with different sizes in the same model family. Similar patterns hold for the much smaller Qwen-2.5-1.5B model (Figure~\ref{subfig:steer-a} and Figure~\ref{subfig:steer-b}) and larger 14B model (Figure~\ref{fig:steer-14b} in Appendix~\ref{apdx:14b}). Steering along the \direction{} direction (to induce the step use in LLMs' internal reasoning) is weaker for smaller models, whereas the results in the \testb{} are comparable.

\section{Ablation study}

\subsection{Threshold of \scorename{} for extracting steering directions}\label{apdx:threshold-ablation}
\looseness=-1
In this section, we ablate the threshold for selecting true-thinking steps when extracting the \direction{} vector. Our goal is to better understand the scale of \scorename{}, that is, how low a score may already indicate decorative thinking and how high a score reflects true thinking. We use steering performance as an indirect probe of how \scorename{} correlates with the internal engagement of steps in reasoning.

\begin{figure}[ht]
    \centering
    \includegraphics[width=0.8\linewidth]{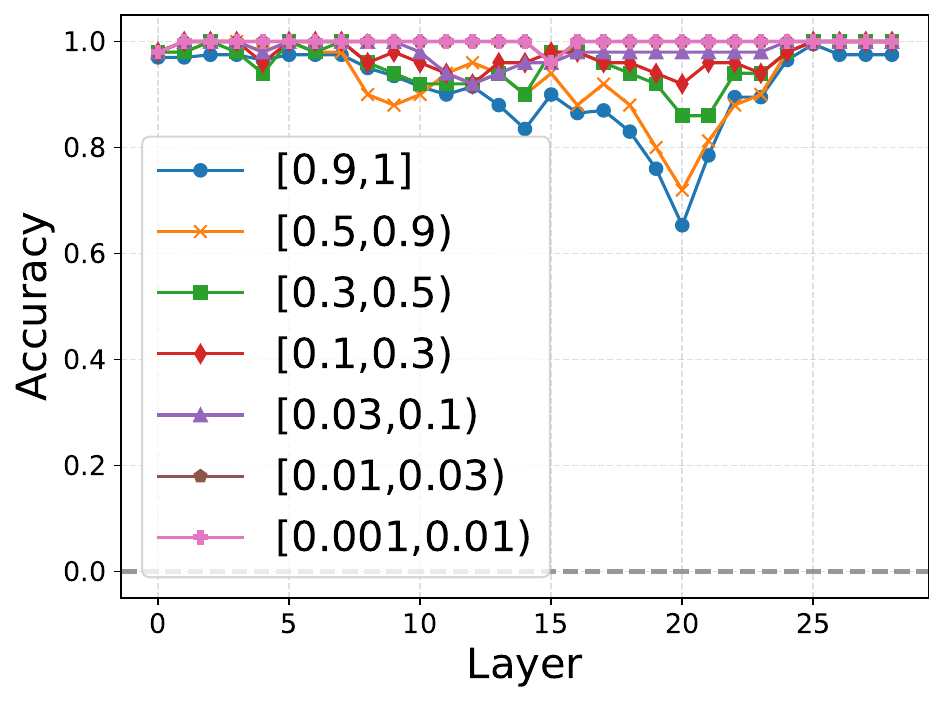}
    \caption{Test results of Nemotron on the \testa{} where \direction{} directions are extracted between examples with zero \scorename{} (as decorative-thinking steps $s_\text{DT}$) and examples of different ranges of \scorename{} (as true-thinking steps $s_\text{TT}$), and the lower accuracy means stronger steering effects. }
    \label{fig:up-threshold}
\end{figure}

\looseness=-1
When extracting steering directions with difference-in-means, the steps with zero \scorename{} are treated as decorative-thinking steps ($s_\text{DT}$), while we use steps from different ranges of \scorename{} as true-thinking steps ($s_\text{TT}$). As shown in Figure~\ref{fig:up-threshold}, when \direction{} directions are computed from steps drawn from increasingly higher \scorename{} ranges, the steering effect grows roughly linearly. In contrast, using steps with \scorename{} below 0.03 yields negligible steering, suggesting these steps are internally treated as decorative, similarly to those with zero \scorename{}. Because the \direction{} directions are computed as the difference in mean hidden states between true and decorative steps (Eq.~\ref{eq:dir}), negligible steering effects mean the direction fails to capture the meaningful distinction between decorative and true thinking.  

Overall, this analysis reveals an implicit decision boundary in the model’s internal space: while the distribution of \scorename{} is very long-tailed and high-\scorename{} steps are rare, larger \scorename{} indeed corresponds to genuinely influential reasoning.  We leave further in-depth study of the geometry~\citep{wollschlager2025the} of hidden states and \direction{} directions in the latent space as future work.

\subsection{Averaging over $\text{ATE}(c)$ for \scorename{}}\label{apdx:ablation-ate}

We compare using only $\text{ATE}(1)$ in \scorename{} with the complete \scorename{} to identify true-thinking steps. This slightly differs from the DropStep method in Table~\ref{tab:ablation-ate}, which is adopted by past works~\citep{tanneru2024hardness,lanham2023measuring,wang-etal-2025-chain,bogdan2025thought}. DropStep defines $\text{ATE}(1)$ with step removal as the perturbation, i.e., $\text{ATE}(1)^{\text{drop}}=\mathrm{P}(y^* \mid \mathcal{C},s)-\mathrm{P}(y^* \mid \mathcal{C})$. In contrast, here $\text{ATE}(1)=\mathrm{P}(y^* \mid \mathcal{C},s)-\mathrm{P}(y^* \mid \mathcal{C},s')$ using numerical perturbation by changing the numbers in step $s$. However, as shown in Table~\ref{tab:ablation-ate}, incorporating $\text{ATE}(0)$ is still necessary, as the resulting \direction{} direction achieves stronger intervention performance.

\begin{table}[t]
  \centering
  \caption{Results on MATH for steering with directions identified from true-thinking steps based on ATE(1) and complete \scorename{}.}
  \label{tab:ablation-ate}
  \small
  \setlength{\tabcolsep}{3pt}
  \renewcommand{\arraystretch}{1.1}
  \begin{tabular}{l S[table-format=2.1] S[table-format=2.1] S[table-format=2.1] S[table-format=2.1]}
    \toprule
    & \multicolumn{2}{c}{\shortstack{$|\mathrm{ATE}_{\text{nec}}(1)|$\\ only}}
    & \multicolumn{2}{c}{\shortstack{$\mathrm{mean}\big(|\mathrm{ATE}_{\text{nec}}(1)|,$\\ $|\mathrm{ATE}_{\text{suf}}(0)|\big)$}} \\
    \cmidrule(lr){2-3}\cmidrule(lr){4-5}
    & {A} & {B} & {A} & {B} \\
    \midrule
    Flip rate & 10.0 & 41.6 & 38.0 & 62.5 \\
    \bottomrule
  \end{tabular}
\end{table}

\begin{figure}[ht]
\centering
\setlength{\fboxsep}{6pt} 
\fcolorbox{black!10}{gray!3}{%
\begin{minipage}{0.9\linewidth}
\footnotesize

\textbf{Problem:}\;
A square has a side length of $10$ inches. Congruent isosceles right triangles are cut off each corner so that the resulting octagon has equal side lengths. How many inches are in the length of one side of the octagon? Express your answer as a decimal to the nearest hundredth.

\textbf{CoT:}\;...Then, the side length of the octagon is $x\sqrt{2} = \frac{10}{\sqrt{2}+2}\cdot\sqrt{2} = \frac{10\sqrt{2}}{\sqrt{2}+2}$, which simplifies to $10\sqrt{2}-10 \approx 4.142$ inches. So, $4.14$ inches.

\begingroup\setlength{\fboxsep}{2pt}%
\colorbox{lightblue}{%
  \parbox{\dimexpr\linewidth-2\fboxsep\relax}{%
  Wait, but let me compute $10\sqrt{2}-10$ more accurately.
  $\sqrt{2}\approx 1.41421356$, so $10\sqrt{2}\approx 14.1421356$ and
  $14.1421356-10=4.1421356$ inches, which rounds to $4.14$ inches.}%
}%
\endgroup

So, the length of one side of the octagon is approximately 4.14.

\end{minipage}}
\caption{ An example of unfaithful self-verification steps (highlighted in blue) where the \scorename{} score of each step is found smaller than 0.005. Low \scorename{} indicates that those steps are not truly engaged in computation; rather, these reasoning steps are likely to be decorative and function as an appearance of self-verification, contributing minimally to the model's final prediction. }
\label{fig:example-self-check-low-score}
\end{figure}

\clearpage
\newpage

\section{Examples with low TTS}\label{apdx:example_low_tts}
We provide more examples where reasoning steps have TTS lower than 0.005. For this section, we classify each reasoning example based on behavioral patterns labeled in \citet{venhoff2025understanding}

\newcommand{\ctxblock}[1]{\begingroup\setlength{\fboxsep}{1ex}\colorbox{black!8}{\parbox{\linewidth}{\footnotesize \textbf{Context steps (last two sentences):}\ #1}}\endgroup}
\newcommand{\stepblock}[2]{\begingroup\setlength{\fboxsep}{1ex}\colorbox{red!12}{\parbox{\linewidth}{\small \textbf{Steps with low TTS --- #1:}\ #2}}\endgroup}

\noindent\textbf{Uncertainty-estimation example}

\ctxblock{Another way is to divide the quadrilateral into two triangles and find the area of each triangle, then add them up. Looking at the points, maybe I can split the quadrilateral along the x-axis}

\stepblock{Uncertainty-estimation}{ The points (-1, 0) and (2, 0) are on the x-axis, so the quadrilateral is split into two triangles: one above the x-axis and one below. 

Wait, actually, the point (0, 1) is above the x-axis, and (0, -3) is below.  So, if I connect (-1, 0) to (0, 1) to (2, 0) to (0, -3) and back to (-1, 0), it's a quadrilateral that crosses over the x-axis. 

Alternatively, maybe I can split it into two triangles: one with vertices (-1, 0), (0, 1), (2, 0) and another with vertices (2, 0), (0, -3), (-1, 0).  Wait, but that might not cover the entire quadrilateral. 

Alternatively, perhaps split it into triangles with a common diagonal.  Let me see. 

If I connect (0, 1) to (0, -3), that's a vertical line along the y-axis.  Then, the quadrilateral is split into two parts: one on the left side of the y-axis and one on the right. 

But actually, the points are (-1, 0), (0, 1), (2, 0), (0, -3)}

\noindent\textbf{Uncertainty-estimation example}

\ctxblock{Looking at the points, maybe I can split the quadrilateral along the x-axis. The points (-1, 0) and (2, 0) are on the x-axis, so the quadrilateral is split into two triangles: one above the x-axis and one below}

\stepblock{Uncertainty-estimation}{

Wait, actually, the point (0, 1) is above the x-axis, and (0, -3) is below.  So, if I connect (-1, 0) to (0, 1) to (2, 0) to (0, -3) and back to (-1, 0), it's a quadrilateral that crosses over the x-axis. 

Alternatively, maybe I can split it into two triangles: one with vertices (-1, 0), (0, 1), (2, 0) and another with vertices (2, 0), (0, -3), (-1, 0).  Wait, but that might not cover the entire quadrilateral. 

Alternatively, perhaps split it into triangles with a common diagonal.  Let me see. 

If I connect (0, 1) to (0, -3), that's a vertical line along the y-axis.  Then, the quadrilateral is split into two parts: one on the left side of the y-axis and one on the right. 

But actually, the points are (-1, 0), (0, 1), (2, 0), (0, -3)}

\noindent\textbf{Backtracking example}

\ctxblock{So, putting it all together, 20\% of 50\% of 80 is 8. That seems correct, but let me verify it another way to make sure I didn't make a mistake}

\stepblock{Backtracking}{

Another approach is to multiply all the percentages together first and then apply them to 80.  So, 20\% is 0. 2, and 50\% is 0. 5.  Multiplying those together: 0. 2 * 0. 5 = 0. 1}


\noindent\textbf{Adding-knowledge example}

\ctxblock{Wait, but that might not be accurate. Alternatively, maybe split the quadrilateral into two triangles by drawing a diagonal from (-1, 0) to (2, 0)}

\stepblock{Adding-knowledge}{ Then, the quadrilateral is split into two triangles: one with vertices (-1, 0), (0, 1), (2, 0) and another with vertices (-1, 0), (2, 0), (0, -3).  Let me calculate the area of each triangle and add them. 

First triangle: (-1, 0), (0, 1), (2, 0)

Using the formula for the area of a triangle with coordinates: 1/2 |x1(y2 - y3) + x2(y3 - y1) + x3(y1 - y2)|

Plugging in the points:

x1 = -1, y1 = 0

x2 = 0, y2 = 1

x3 = 2, y3 = 0

So,

Area = 1/2 | (-1)(1 - 0) + 0(0 - 0) + 2(0 - 1) |

= 1/2 | (-1)(1) + 0 + 2(-1) |

= 1/2 | -1 + 0 - 2 |

= 1/2 | -3 | = 1/2 * 3 = 1}

\noindent\textbf{Adding-knowledge example}

\ctxblock{Let me calculate the area of each triangle and add them. First triangle: (-1, 0), (0, 1), (2, 0)

Using the formula for the area of a triangle with coordinates: 1/2 |x1(y2 - y3) + x2(y3 - y1) + x3(y1 - y2)|

Plugging in the points:

x1 = -1, y1 = 0

x2 = 0, y2 = 1

x3 = 2, y3 = 0

So,

Area = 1/2 | (-1)(1 - 0) + 0(0 - 0) + 2(0 - 1) |

= 1/2 | (-1)(1) + 0 + 2(-1) |

= 1/2 | -1 + 0 - 2 |

= 1/2 | -3 | = 1/2 * 3 = 1.5}

\stepblock{Adding-knowledge}{
Second triangle: (-1, 0), (2, 0), (0, -3)

Again, using the same formula:

x1 = -1, y1 = 0

x2 = 2, y2 = 0

x3 = 0, y3 = -3

Area = 1/2 | (-1)(0 - (-3)) + 2((-3) - 0) + 0(0 - 0) |

= 1/2 | (-1)(3) + 2(-3) + 0 |

= 1/2 | -3 -6 + 0 |

= 1/2 | -9 | = 1/2 * 9 = 4. 5

Adding both areas: 1. 5 + 4. 5 = 6

Okay, so that's the same result as before}

\end{document}